\documentclass{article}

\usepackage[preprint]{neurips_2025}

\usepackage{tabularx}
\usepackage[most]{tcolorbox}   
\usepackage{xcolor}            

\usepackage{amsthm}
\usepackage{mdframed}

\usepackage{amsopn}

\usepackage{amsthm}
\usepackage{amsmath}
\usepackage{amsopn}
\usepackage{subcaption}

\theoremstyle{plain}

\theoremstyle{definition}

\definecolor{color1}{rgb}{0.1,0.7,0.8} 
\definecolor{color2}{rgb}{0.9,0.1,0.1} 
\definecolor{color3}{rgb}{0.7,0.3,0.7} 
\definecolor{color4}{rgb}{0.3,0.3,0.7} 
\definecolor{color5}{RGB}{8, 102, 3} 
\definecolor{color6}{rgb}{0.53, 0.66, 0.42} 

\usepackage{amsthm}

\usepackage{amsopn}

\usepackage{hyperref}
\usepackage{amsmath}
\usepackage[utf8]{inputenc} 
\usepackage[T1]{fontenc}    
\usepackage{hyperref}       
\usepackage{url}            
\usepackage{booktabs}       
\usepackage{amsfonts}       
\usepackage{nicefrac}       
\usepackage{microtype}      
\usepackage{xcolor}         

\usepackage{algorithm}          
\usepackage{algpseudocode}      
\usepackage{graphicx}

\usepackage{multirow}
\usepackage{graphicx}   
\usepackage{booktabs}   
\usepackage{enumitem}   
\usepackage{amsmath}    
\usepackage{amssymb}    

\usepackage{authblk}

\usepackage{amsmath}
\usepackage{graphicx}
\usepackage{tcolorbox}
\usepackage{booktabs}
\usepackage{xcolor}
\usepackage{pgfplots}
\usepackage{listings}
\usepackage{caption}
\usepackage{mdframed}
\usepackage{enumitem}

\pgfplotsset{compat=1.16}

\newcounter{scenariocount}
\newenvironment{scenario}[1]{%
  \refstepcounter{scenariocount}%
  \begin{tcolorbox}[colback=gray!5,colframe=gray!40,title=Scenario \thescenariocount: #1]
}{%
  \end{tcolorbox}
}

\newenvironment{traitprofile}[1][]{%
  \begin{tcolorbox}[colback=blue!5,colframe=blue!30,title=#1]
}{%
  \end{tcolorbox}
}

\usepackage{tikz}                     
\usepackage{pgfplots}                 
\pgfplotsset{compat=1.18}             

\usepackage{tikz}
\usepackage{pgfplots}
\pgfplotsset{compat=1.18}   
\usepackage{xcolor}         

\usepackage{tikz}
\usepackage{xcolor}

\usepackage{float}

\usepackage{amsthm}

\theoremstyle{plain}

\theoremstyle{definition}

\theoremstyle{remark}

\usepackage{amsopn}               

\usepackage[utf8]{inputenc}
\usepackage[T1]{fontenc}
\usepackage{hyperref}
\usepackage{url}
\usepackage{booktabs}
\usepackage{nicefrac}
\usepackage{microtype}
\usepackage{xcolor}
\usepackage{tabularx}
\usepackage{graphicx}
\usepackage{multirow}
\usepackage{enumitem}
\usepackage{algorithm}
\usepackage{algpseudocode}

\usepackage{tikz}
\usetikzlibrary{shapes.geometric, arrows, shadows, backgrounds, positioning}
\usepackage{xcolor}

\definecolor{headerBlue}{HTML}{2563EB}
\definecolor{darkBlue}{HTML}{1E40AF}
\definecolor{userGray}{HTML}{94A3B8}
\definecolor{darkGray}{HTML}{64748B}
\definecolor{scenarioBlue}{HTML}{3B82F6}
\definecolor{analysisBlue}{HTML}{1D4ED8}
\definecolor{traitPurple}{HTML}{7C3AED}
\definecolor{confidencePurple}{HTML}{8B5CF6}
\definecolor{decisionYellow}{HTML}{F59E0B}
\definecolor{profileGreen}{HTML}{059669}
\definecolor{agentRed}{HTML}{E11D48}
\definecolor{specGray}{HTML}{334155}
\definecolor{bgGray}{HTML}{F8FAFC}
\definecolor{lineGray}{HTML}{E2E8F0}

\usetikzlibrary{arrows.meta,positioning,shapes,fit,backgrounds,calc,shadows,decorations.pathreplacing}

\definecolor{modulebg}{RGB}{245,245,255}
\definecolor{processcolor}{RGB}{230,230,250}
\definecolor{memorycolor}{RGB}{230,250,230}
\definecolor{emotioncolor}{RGB}{255,230,230}
\definecolor{planningcolor}{RGB}{255,240,220}
\definecolor{socialcolor}{RGB}{220,240,255}
\definecolor{goalcolor}{RGB}{250,230,250}
\definecolor{globalcolor}{RGB}{255,255,220}
\definecolor{inputcolor}{RGB}{200,230,255}
\definecolor{outputcolor}{RGB}{200,255,200}

\usepackage{tikz}
\usepackage{xcolor}
\usepackage{float}

\usetikzlibrary{shapes,arrows,positioning,fit,backgrounds,calc,shapes.geometric,shapes.misc}

\usepackage{caption}

\title{CogniPair: From LLM Chatbots to Conscious AI Agents - GNWT-Based Multi-Agent Digital Twins for Social Pairing -  Dating \& Hiring Applications}

\usepackage[preprint]{neurips_2025}

\usepackage{authblk}

\title{CogniPair: From LLM Chatbots to Conscious AI Agents - GNWT-Based Multi-Agent Digital Twins for Social Pairing - Dating \& Hiring Applications}

\author{
Wanghao Ye$^1$, 
Sihan Chen$^2$, 
Yiting Wang$^1$, 
Shwai He$^1$, 
Bowei Tian$^1$, 
Guoheng Sun$^1$, 
Ziyi Wang$^1$, 
Ziyao Wang$^1$, 
Yexiao He$^1$, 
Zheyu Shen$^1$, 
Meng Liu$^1$, 
Yuning Zhang$^1$, 
Meng Feng$^1$, 
Yang Wang$^3$, 
Siyuan Peng$^1$, 
Yilong Dai$^4$, 
Zhenle Duan$^1$, 
Lang Xiong$^1$,
Joshua Liu$^1$,
Hanzhang Qin$^{5,*}$, 
Ang Li$^{1,*}$
}

\begin{document}

\maketitle
\vspace{-3 em}
\begin{center}
\small
$^1$University of Maryland, College Park\\
$^2$Viterbi School of Engineering, University of Southern California\\
$^3$Northeastern University\\
$^4$Department of Computer Science, University of Florida\\
$^5$National University of Singapore\\
$^*$Joint corresponding authors

\vspace{0.3em}
\scriptsize
\textbf{University of Maryland Departments:} Electrical and Computer Engineering, Computer Science, \\
Robert H. Smith School of Business, School of Public Policy, Department of Agricultural and Resource Economics

\vspace{0.3em}
\{wy891, yiting, bowei, guoheng, ziyi, ziyao, yexiao, zheyu, meng, yuning, mfeng, siyuan, zhenle, angli\}@umd.edu\\
sihan@usc.edu \quad yang@northeastern.edu \quad yilong@ufl.edu \quad hanzhang@nus.edu.sg
\end{center}

\begin{abstract}
Current large language model agents lack authentic human psychological processes necessary for genuine digital twins. We present the first computational implementation of Global Workspace Theory (GNWT), creating agents with multiple specialized sub-agents (emotion, memory, social norms, planning, goal-tracking) coordinated through a global workspace broadcast mechanism. This architecture allows agents to maintain consistent personalities while evolving through social interaction. Our CogniPair simulation platform deploys 551 GNWT-Agents for speed dating interactions, grounded in real data from the Columbia University Speed Dating dataset. Evaluations show unprecedented psychological realism, with agents achieving 72\% correlation with human attraction patterns and outperforming baselines in partner preference evolution (72.5\% vs. 61.3\%). Human validation studies confirm our approach's fidelity, with participants rating their digital twins' behavioral accuracy at 5.6/7.0 and agreeing with their choices 74\% of the time. This work establishes new benchmarks for psychological authenticity in AI systems and provides a foundation for developing truly human-like digital agents.
\end{abstract}

\section{Introduction}
Human social interactions—from dating to job interviews—require not just coherent dialogue but authentic psychological processes including emotion regulation, memory consolidation, and dynamic preference formation. LLM-based agents have been applied to model human social interactions, showing promise in domains such as customer service, healthcare assistance, and educational tutoring~\cite{Park2023, Xu2023, Wang2024}. Despite recent advances, current LLM-based agents face two fundamental limitations that restrict their ability to model human behavior realistically: (1) the \textbf{psychological behavior gap}—they cannot authentically simulate internal mental states, emotional processing, or evolving preferences~\cite{Zhang2024b, Serapio-Garcia2023, Jiang2023,Guo2024}; and (2) the \textbf{social behavior gap}—they fail to capture the complex dynamics of human-to-human interactions where preferences and behaviors co-evolve through social experiences~\cite{Park2023, Zhang2018, Huang2024, Park2022social, Aher2023, Sun2024}.

The \textbf{psychological behavior gap} manifests in two critical problems: the \textbf{individualization problem}, where agents act like generic humans rather than specific individuals with unique psychological profiles, and the \textbf{static personality problem}, where agents cannot evolve mentally through experience. Existing approaches such as Stanford's Generative Agents~\cite{Park2023} demonstrated emergent behaviors but relied on fictional personas without real human data. PersonaChat~\cite{Zhang2018} introduced personality descriptions that remain synthetic and fixed. Recent personality modeling efforts~\cite{Serapio-Garcia2023, Jiang2023} achieve only surface-level behavioral mimicry without cognitive grounding. Most critically, these approaches treat personality as immutable prompts rather than dynamic psychological states shaped by experience.

The \textbf{social behavior gap} emerges when attempting to model authentic social interactions, particularly in complex domains such as relationship formation. Current LLM agents lack the capability to engage in authentic social dynamics where preferences evolve through interaction, emotional responses adapt to social feedback, and behavioral patterns shift based on interpersonal experiences. This limitation becomes particularly pronounced in domains requiring complex social cognition, such as dating scenarios where mutual attraction emerges through dynamic, bidirectional assessment processes.

\begin{figure}[h]
\centering
\includegraphics[width=0.67\linewidth]{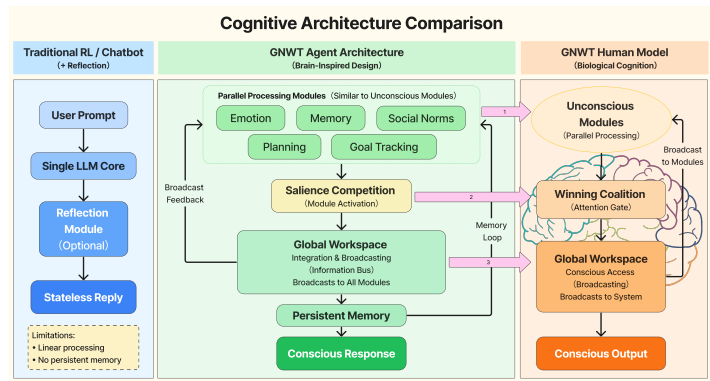}
\caption{ Comparison of Cognitive Architectures: Traditional RL/Chatbot (left), GNWT Agent Architecture (center), and GNWT Human Model (right), showing the evolution from linear processing to brain-inspired parallel processing with global workspace integration.}
\label{fig:cognitive-architectures-comparison}
\end{figure}

To address these fundamental gaps, we turn to Global Workspace Theory (GNWT)~\cite{Mashour2020, Baars2005}, a leading neurocognitive model that explains how human consciousness emerges from the interaction of specialized brain modules. In human cognition, GNWT describes how disparate neural processes—emotion, memory, perception, and planning—compete for access to a central "global workspace." When information becomes sufficiently salient, it triggers a broadcast that propagates this content throughout the brain, creating our unified stream of consciousness. This theoretical foundation provides a clear roadmap for building agents that can overcome both the psychological and social behavior gaps(Figure~\ref{fig:cognitive-architectures-comparison}).
To address the \textbf{psychological behavior gap}, we operationalize GNWT into a computational agent architecture where each individual agent contains multiple specialized sub-agents working in parallel as a unified consciousness. Our GNWT-Agent implements five specialized cognitive modules—Emotion, Memory, Planning, SocialNorms, and GoalTracking—each grounded in neurocognitive theories and parameterized by the agent's Five-Factor personality profile. By implementing GNWT's broadcast mechanism computationally, we create agents with genuine internal psychological dynamics—emotion sub-agents generate affective responses, memory sub-agents consolidate experiences, social norms sub-agents manage cultural awareness, planning sub-agents develop strategies, and goal-tracking sub-agents maintain objectives. This architecture fundamentally differs from traditional LLM agents that process inputs sequentially without internal state evolution.

To bridge the \textbf{social behavior gap}, we developed \textbf{CogniPair}, a social-influence decision system that enables GNWT-Agents to engage in authentic social interactions and evolve through experience. CogniPair is not merely a testbed but a comprehensive system for modeling and guiding social influence between individuals, ultimately optimizing decision processes across various social contexts. While our primary evaluation uses a speed dating testbed, the CogniPair system itself can be extended to other social decision environments such as team formation, negotiation scenarios, and collaborative problem-solving.

We selected speed dating as our evaluation domain because it exemplifies the most challenging aspects of human social cognition—rapid compatibility assessment, dynamic preference formation, emotional regulation under uncertainty, and integration of multiple information streams. The Columbia University Speed Dating dataset~\cite{Fisman2006, Fisman2008} provides rich behavioral ground truth, including pre- and post-interaction preferences, attraction ratings, and decision outcomes, enabling rigorous evaluation of social realism. Our dating testbed deploys 551 GNWT-Agents in a two-level simulation architecture: the internal level models psychological processes within each agent, while the external level simulates social dynamics between agents. This scale—20 times larger than previous personality-based simulations~\cite{Park2023}—enables statistically valid analysis of emergent social patterns, creates sufficient diversity for complex relationship networks, and allows measurement of population-level phenomena.

Through this dual approach—addressing the psychological gap with GNWT-Agents and the social gap with the CogniPair system—our framework uniquely enables both individualization and dynamic evolution. The global workspace mechanism naturally handles the stability-plasticity dilemma in personality modeling: core traits remain stable through persistent attention patterns while allowing adaptive changes through experience-driven broadcast priority shifts. Unlike previous approaches that treat personality as static prompts or surface behaviors, our architecture models the cognitive processes underlying personality and the social dynamics shaping its evolution.

Our evaluation framework measures psychological and social realism across multiple dimensions. For psychological realism, we assess preference consistency (how well agents maintain core values while adapting), emotional coherence (whether affective responses follow human psychological principles), and memory integration (how past experiences influence decisions). For social realism, we measure attraction correlation (how closely agent mate selection aligns with human patterns), interaction dynamics (how conversations evolve), and emergent social phenomena (group formation, preference shifts). These metrics ground our claims about achieving human-level authenticity in both psychological processing and social behavior.
Our experiments demonstrate that our GNWT-Agents achieve unprecedented accuracy in modeling human social dynamics. We show significant improvements in partner preference evolution (72.5\% accuracy vs. 61.3\% for Multi-Agent Debate), self-perception adaptation, external evaluation changes, and match prediction (77.8\% accuracy) compared to state-of-the-art baselines. Most impressively, our agents demonstrate human-like evolution patterns with high correlation to ground truth data (above 0.7 across multiple dimensions, with a 0.72 correlation for match patterns). Human validation studies further confirm the psychological fidelity of our approach, with participants rating their digital twins' behavioral fidelity at 5.6/7.0 and agreeing with their twin's choices 74\% of the time. Our key contributions are:
\begin{enumerate}
\itemsep=0pt
\item We are the \textbf{first to operationalize GNWT for computational agents}, creating a cognitive architecture where multiple sub-agents within each agent replicate human psychological processes through dynamic workspace broadcasting
\item We develop CogniPair, the \textbf{first social-influence decision system combining cognitive theory with large-scale social simulation}, capable of generalizing beyond our dating testbed to various social decision environments as demonstrated by successful transfer to job interview contexts (81\% accuracy)
\item Our extensive experiments confirm GNWT's broadcast mechanism enables \textbf{genuine personality evolution}, with significant improvements in partner preference evolution (72.5\% vs. 61.3\%), self-perception adaptation, and external evaluation shifts compared to state-of-the-art baselines
\end{enumerate}

\section{Related Work}
\textbf{LLMs for Social Interaction and Simulation: } Recent LLM advances have enabled sophisticated conversational capabilities~\cite{Thoppilan2022,Park2023}, but standard LLMs lack persistent psychological states~\cite{Guo2024,Zhang2024b}. While approaches like Chain-of-Thought~\cite{Wei2022}, self-consistency~\cite{Wang2022Self}, retrieval-augmentation~\cite{lewis2020retrieval}, and memory architectures~\cite{Guo2024,Zhang2024b} enhance reasoning, they show limitations in modeling social dynamics~\cite{Xi2023,Wang2024} and rarely incorporate selective attentional mechanisms~\cite{Zhang2024b,Zhu2023,Andreas2022}.

Social simulation systems like Park et al.'s Generative Agents~\cite{Park2023} implement memory and planning but use fictional personas without psychological grounding. Huang et al.~\cite{Huang2024} identified gaps between simulated and human behavior in LLM agents. While PersonaChat~\cite{Zhang2018}, Li et al.~\cite{Li2024}, and Gao et al.~\cite{Gao2023} advanced social simulation, none incorporate real human data for initialization, neuroscience-based cognitive processing, dynamic personality evolution, or comprehensive social evaluation metrics~\cite{Aher2023,Park2023}. Our CogniPair system integrates all these components to address these fundamental limitations.

\textbf{Modeling Psychological Processes in AI: } Traditional cognitive architectures relied on hand-crafted symbolic representations with limited adaptability. Global Neuronal Workspace Theory (GNWT)~\cite{Baars2005,Mashour2020} provides a framework for modeling human-like dynamic attention allocation. While Bengio et al.~\cite{Bengio2017} and Goyal et al.~\cite{Goyal2022} explored computational implementations of consciousness theories, they focused on perceptual processes rather than higher-order social cognition. Frameworks from Dehaene et al.~\cite{Dehaene2017} and Mashour et al.~\cite{Mashour2020} similarly neglect social-cognitive dimensions essential for authentic human interaction simulation. Recent digital twins research emphasizes behavioral mimicry~\cite{Park2024} without capturing underlying psychological dynamics, and personality modeling systems~\cite{Li2023,wang2023rolellm} typically treat traits as static rather than adaptive characteristics evolving through social interaction~\cite{Acerbi2023,Sun2024}.

Systems using debate mechanisms~\cite{Du2023,chan2023chateval} or transformer-based aggregation~\cite{chen2023agentverse,Hong2023} show improved performance over single-agent approaches but implement coordination through explicit turn-taking rather than human-like parallel processing. Our GNWT-Agent addresses these limitations through a global workspace mechanism enabling selective attention based on personality-driven priorities. CogniPair enables realistic preference adaptation while maintaining consistent core traits, using the Columbia dataset~\cite{Fisman2006} for both initialization and evaluation, demonstrating significantly higher contextual coherence and emotional alignment compared to alternatives.

\begin{figure}
    \centering
    \includegraphics[width=0.7\linewidth]{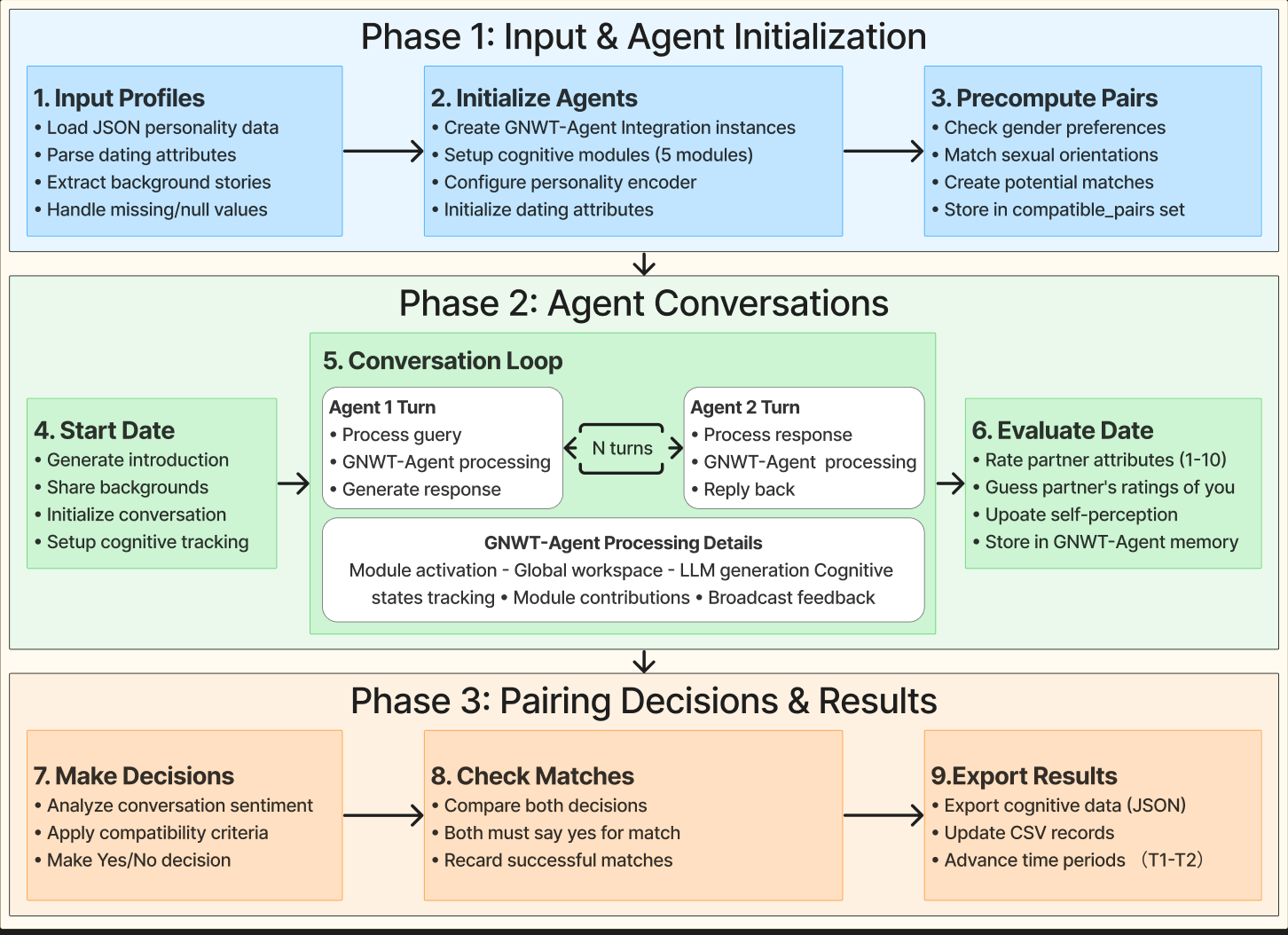}
    \caption{CogniPair Platform System Flow}
    \label{fig:cognipair_system_flow}
    \vspace{-10pt}
\end{figure}

\section{CogniPair: Cognitive Social Pairing Agent System}

In this section, we introduce single GWNT-Agent's structure (Sec. 3.1),  how the single-turn conversation generated from GNWT-Agent (Sec. 3.2), and simulated social environment that allows multi-turn conversations (Sec. 3.3),

\subsection{GNWT-Agent Cognitive Modules}

\textbf{Core Cognitive Architecture:} GNWT-Agent's cognitive processing is based on the Global Neuronal Workspace Theory (GNWT), which provides a computational model of human consciousness and cognitive processing mechanisms (Algorithm~\ref{alg:cognipair}, Figure~\ref{fig:GNWT-Agent_architecture}). In implementation, we deploy five competing cognitive modules, each focusing on different cognitive functional domains(detailed system flow in Appendix~\ref{appendix:gnwt_agent})):

\textbf{Emotion Module}. This module performs in three stages: (1) emotion detection, identifying affective markers in text through feature extraction; (2) valence-arousal assessment, mapping detected emotions to a two-dimensional affective space; and (3) regulation strategy generation, adjusting emotional response intensity based on the agent's neuroticism ($N$) parameter. Mathematically represented as: $R_{\text{Emotion}} = f_{\text{E}}(Q, H, \mathrm{GW}, N)$, where higher $N$ values amplify emotional processing weights.

\textbf{Memory Module}. This module maintains a dual memory system: (1) episodic memory, storing time-stamped dialogue segments and interaction patterns; and (2) semantic memory, preserving knowledge about conversation topics and abstract concepts. The retrieval process employs vector similarity search, with the openness parameter ($O$) adjusting memory breadth and retrieval strategies: $R_{\text{Memory}} = f_{\text{M}}(Q, H, \mathrm{GW}, O)$.

\textbf{Planning Module}. This module implements hierarchical goal decomposition: breaking down complex social goals (e.g., "establish rapport") into tactical steps (e.g., "identify common interests," "express empathy"). This process is regulated by the conscientiousness parameter ($C$), which controls planning depth and strategic rigor: $R_{\text{Planning}} = f_{\text{P}}(Q, H, \mathrm{GW}, C)$.

\textbf{SocialNorms Module}. This module maintains a knowledge base of social interaction rules, evaluating the appropriateness of conversational behaviors. Processing includes: (1) etiquette checking, verifying response politeness; (2) boundary monitoring, preventing excessive self-disclosure; and (3) reciprocity verification, ensuring balanced conversational contributions. The agreeableness parameter ($A$) adjusts the strictness of norm enforcement: $R_{\text{SocialNorms}} = f_{\text{SN}}(Q, H, \mathrm{GW}, A)$.

\textbf{GoalTracking Module}. This module continuously evaluates conversational progress: (1) direction monitoring, tracking advancement toward preset objectives; (2) uncertainty assessment, identifying information gaps that require clarification; and (3) direction adjustment, recalibrating goals based on interaction dynamics. The extraversion parameter ($E$) influences goal assertiveness: $R_{\text{GoalTracking}} = f_{\text{GT}}(Q, H, \mathrm{GW}, E)$.

\subsection{Single Decision-Making System Flow}

The CogniPair system implements a structured decision-making flow that systematically processes social interactions through nine distinct stages across three operational phases, as illustrated in Figure~\ref{fig:cognipair_system_flow} \& Algorithm~\ref{alg:dating_simulation}.

\textbf{Phase 1: Input \& Agent Initialization} establishes the foundation for social simulations. In Step 1 (\textit{Input Profiles}), the system ingests personality profiles, preference distributions, and demographic data, either from human participant records (e.g., the Columbia Speed Dating dataset) or synthetically generated profiles with balanced demographic distributions. Step 2 (\textit{Initialize Agents}) instantiates GNWT-Agent cognitive architectures for each participant, mapping Five-Factor personality traits to module weights and initializing the global workspace with prior knowledge. This process follows $\text{Agent}_i = \textsc{InitializeAgent}(\text{Profile}_i, \theta_{\text{module}})$, where $\theta_{\text{module}}$ represents the module-specific parameters. In Step 3 (\textit{Precompute Pairs}), the system generates potential interaction dyads based on specified criteria (e.g., gender preferences, age constraints), creating a pairing matrix $P_{m\times n}$ where each element $p_{ij}$ indicates pairing eligibility between agents $i$ and $j$.

\textbf{Phase 2: Agent Conversations} executes the multi-turn interactive dialogues. Step 4 (\textit{Start Session}) initializes the conversational context $C_0$ with environmental parameters (e.g., spatial configuration, temporal constraints) and interaction goals. Step 5 (\textit{Conversation Loop}) implements the turn-taking dynamics where each agent processes inputs through their cognitive modules and generates responses. For each turn $t$, Agent 1 generates response $R_{1,t} = \textsc{ProcessInput}(Q_t, H_{t-1}, \text{GW}_1, P_1)$ where $Q_t$ is the current query, $H_{t-1}$ is the conversation history, $\text{GW}_1$ is Agent 1's global workspace state, and $P_1$ represents personality parameters. 

Within each processing cycle, the agent integrates outputs from all cognitive modules to form a coherent response through the $\textsc{IntegrateModuleOutputs}$ function (Algorithm~\ref{alg:cognipair} Line 29), formalized as a personality-weighted combination:
\begin{align}
\text{Response} &= \textsc{IntegrateModuleOutputs}(\{R_M\}, \mathrm{GW}, P) \\
&= \sum_{M \in \text{Modules}} \alpha_M(P) \cdot R_M + \beta(P) \cdot G(\mathrm{GW})
\end{align}
where $\alpha_M(P)$ represents personality-based module weights, $\beta(P)$ is the integration coefficient for global workspace content, and $G(\mathrm{GW})$ extracts key content from the global workspace. This combination strategy ensures the final response reflects both specialized processing from each module and maintains global coherence.

Agent 2 follows an identical process, creating a bidirectional exchange repeated for $N$ turns. Each GNWT-Agent processing instance involves all five cognitive modules competing for global workspace access, with broadcasts occurring when salience exceeds the threshold $\tau$. During processing, emotional reactions, memory retrieval, planning strategies, social norm evaluations, and goal assessments are computed in parallel, with integration weighted by personality parameters.

In Step 6 (\textit{Evaluate Session}), agents assess the interaction quality through multiple dimensions: $E_i = \{E_{\text{attr}}, E_{\text{similar}}, E_{\text{comfort}}, E_{\text{interest}}\}$, with these evaluations stored in the Memory module for subsequent retrieval. The system's adaptive learning mechanism is implemented through two key update functions: (1) the $\text{Memory}.\textsc{updateLongTerm}(Q, \text{Response}, H)$ function (Algorithm~\ref{alg:cognipair} Line 30) stores the current interaction in long-term memory, using attention-based memory consolidation techniques that highlight emotionally salient and goal-relevant content; and (2) the $\textsc{UpdatePreferences}(P, Q, \text{Response}, H)$ function (Algorithm~\ref{alg:cognipair} Line 31) adjusts personality weights based on interaction experiences, implementing fine-tuning learning:
\begin{align}
P_{t+1} = P_t + \eta \cdot \nabla_{P}J(P_t, Q, \text{Response}, H)
\end{align}
where $\eta$ is a learning rate parameter and $\nabla_{P}J$ is the gradient of an objective function measuring interaction success with respect to personality parameters. This dual update mechanism enables agents to continuously evolve their preferences and behaviors based on accumulated social experiences.

\textbf{Phase 3: Pairing Decisions \& Results} culminates in match determinations. In Step 7 (\textit{Make Decisions}), each agent formulates a binary decision (accept/reject) regarding potential future interactions: $D_i = \textsc{DecisionFunction}(E_i, P_i, H, \text{GW}_i)$, where the decision function integrates evaluation metrics, personality preferences, and interaction history. Step 8 (\textit{Check Matches}) identifies mutual matches where both agents express interest: $M_{ij} = D_i \land D_j$, creating a symmetric match matrix. Finally, Step 9 (\textit{Export Results}) aggregates and formats simulation outcomes, including match decisions, preference evolutions, perception changes, and interaction quality metrics, generating comprehensive datasets for subsequent analysis.

This workflow implements the three complexity levels described earlier: low-complexity interactions utilize direct module selection, moderate-complexity interactions employ iterative processing with conflict resolution, and high-complexity interactions integrate the complete multi-phase protocol with comprehensive state tracking. CogniPair's decision-making and learning mechanisms are deliberately scenario-agnostic, allowing application across diverse social contexts—from optimizing information exchange in professional settings to fostering emotional connection in personal relationships. Through this systematic approach, CogniPair captures both the cognitive micromechanics of individual decision-making and the emergent macropatterns of social pairing dynamics.

\subsection{Simulated Social Environment Setup }

\subsubsection{Generalized Environment Parameterization}
CogniPair implements a flexible parameterization system for modeling diverse social interaction environments. The system encapsulates interaction contexts through a comprehensive parameter space $C$ defined as:
\vspace{-2pt}
\begin{align}
C &= \{\text{physical}_p, \text{temporal}_t, \text{social}_s, \text{cultural}_c\} \\
\text{physical}_p &= \{\text{spatial\_layout}, \text{proximity}, \text{sensory\_conditions}\} \\
\text{temporal}_t &= \{\text{duration}, \text{pacing}, \text{sequence\_structure}\} \\
\text{social}_s &= \{\text{group\_size}, \text{relationship\_dynamics}, \text{power\_structure}\} \\
\text{cultural}_c &= \{\text{normative\_expectations}, \text{communication\_styles}\}
\end{align}

The physical parameters capture environmental conditions including spatial arrangements, interpersonal distance, and sensory factors (lighting, acoustics, temperature) that influence interaction dynamics. Temporal parameters define interaction timeframes, turn-taking pacing, and structural sequencing that shape conversational flow. Social parameters model group composition, pre-existing relationship dynamics, and authority structures. Cultural parameters encode normative behaviors and communication conventions appropriate to specific contexts.

This generalized parameterization enables CogniPair to simulate diverse interaction scenarios—from professional meetings to casual gatherings, educational exchanges to intimate conversations—by appropriately configuring these parameters. The system generates contextually-appropriate prompts using natural language templates that translate numerical parameter values into detailed environmental descriptions accessible to language models, enhancing  validity across different simulation contexts.

\subsubsection{Multi-Agent Interaction Architecture}
CogniPair's interaction architecture (Algorithm~\ref{alg:cognipair}) provides a flexible framework for simulating multi-agent social dynamics across diverse scenarios. The system initializes a pool $\mathcal{A}$ of agents, each equipped with five cognitive modules and a global workspace. The initialization process maps individual agent characteristics into module weights and interaction preferences:
\vspace{-1pt}
\begin{align}
\mathrm{IA}_i &= \text{InitializeInteractionAttributes}(P_i) \\
&= \{w_1, w_2, ..., w_n\} \\
\mathrm{Modules}_i.\text{weights} &= M(P_i, \mathrm{IA}_i)
\end{align}
\vspace{-1pt}
where $P_i$ represents the agent's personality profile (typically Five-Factor traits), $\mathrm{IA}_i$ represents scenario-specific interaction attributes, and $M$ is a mapping function that determines module processing parameters. This approach ensures individual agents retain consistent core traits while adapting their behavior appropriately to different social contexts.

The interaction protocol supports multiple engagement patterns:
1. Dyadic exchanges - One-to-one interactions with reciprocal turn-taking; 
2. Group discussions - Multi-participant exchanges with dynamic speaker selection; 
3. Hierarchical interactions - Structured exchanges with defined role-based communication paths.

For each interaction, the system manages turn-taking, tracks interaction histories, and computes evolving relationship metrics.  The architecture records comprehensive interaction data including:
- Complete interaction histories $\mathcal{H}$
- Cognitive trace datasets $\mathcal{T}$ capturing internal mental states
- Relationship development trajectories $\mathcal{R}$ tracking interpersonal dynamics
- Emergent social network structures $\mathcal{N}$ documenting group formation.

This multi-level data collection enables both micro-analysis of individual cognitive processes and macro-analysis of emergent social patterns, providing a foundation for validating the system's fidelity to human social behavior across different interaction contexts.

\section{Experiments and Results}
\label{sec:experiments}

\subsection{Experimental Setup}
\label{subsec:experimental_setup}

\label{subsubsec:dataset}

\textbf{Dataset and Simulation Protocol: }The Columbia University Speed Dating dataset ~\cite{Fisman2006} contains records of 551 participants who engaged in 5,500+ four-minute speed dates over 21 sessions, resulting in over 8,300 observations. Each record includes pre-dating attribute self-ratings (1-10 scale), attribute importance ratings (distributing 100 points across 6 attributes), post-dating partner ratings on the same attributes, and match decisions (yes/no interest in seeing a partner again). The six key attributes measured are: attractiveness, sincerity, intelligence, fun, ambition, and shared interests. We instantiate 551 GNWT-Agents as digital twins of the original participants, initializing each agent's personality profile with the Five-Factor traits inferred from participants' self-ratings and importance distributions. The physical and temporal parameters in the CogniPair system are configured to match the original study's environment (bar-restaurant setting, 4-minute interaction, 8-10 conversation turns). For each simulated date, agents engage in 8 conversation turns, after which they update their self-ratings based on interaction experience, rate their partners on the six attributes, and make match decisions. We compare match patterns, preference evolution, and conversation dynamics against ground truth data from the original study.

\textbf{Baselines:} We compare against state-of-the-art approaches: Single Sequential LLM (standard prompt-based approach without specialized cognitive modules), Memory-Enhanced LLM (incorporates retrieval-augmented context) \cite{lewis2020retrieval}, Multi-Agent Debate (simulates internal deliberation through multiple agents) \cite{chan2023chateval}, and Hierarchical Architecture (uses a command structure to organize decision-making) \cite{Du2023}. All baselines use identical data initialization to ensure fair comparison.

\subsection{Results and Key Findings}
\label{subsec:macro_analysis}
We examine population-level social dynamics to assess how well our system bridges the social behavior gap identified in our introduction. Our experimental approach creates digital twins of the Columbia Speed Dating study participants and compares their evolution with the ground truth human data across multiple time points. 

\subsubsection{Evolution of Social Dynamics}
\label{subsubsec:social_dynamics_evolution}

\begin{figure}
    \centering
    \includegraphics[width=0.9\linewidth]{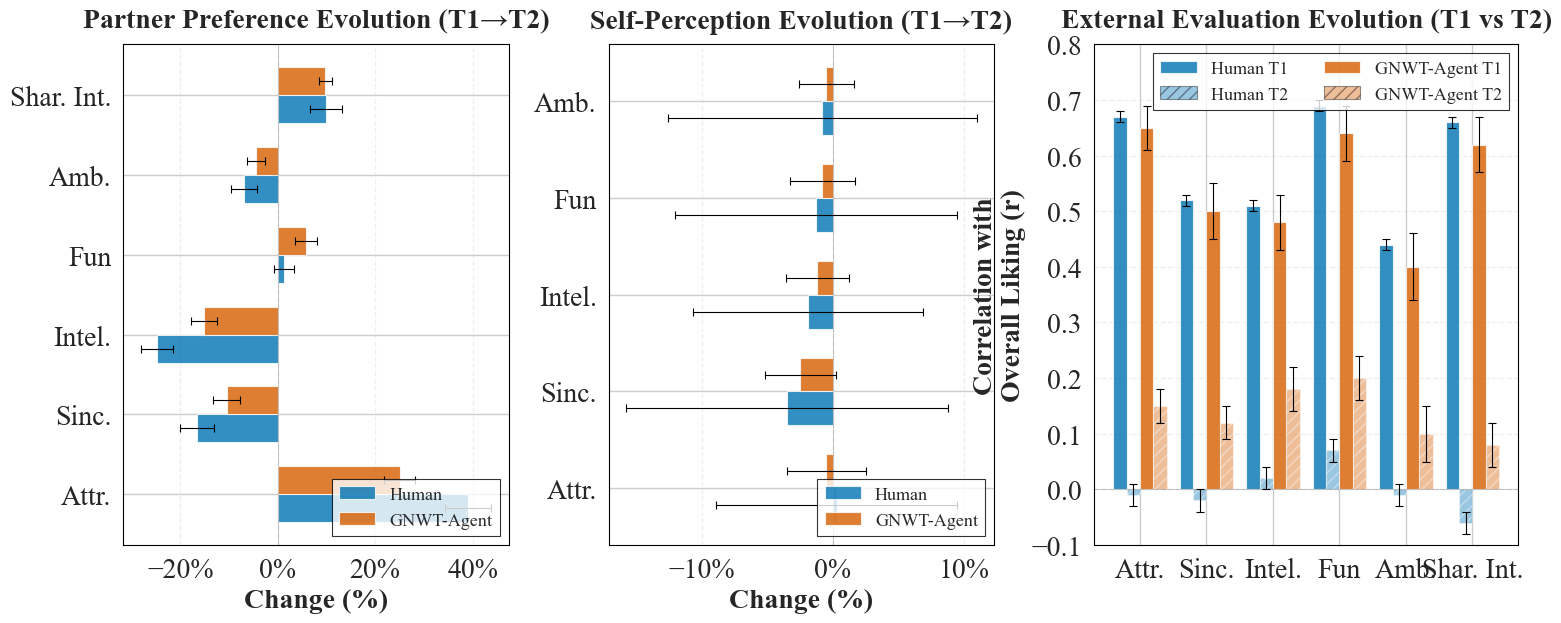}
    \caption{Social Dynamics Evolution: Human vs. GNWT-Agent Comparison. (A) Partner preference changes from T1 to T2; (B) Self-perception adjustments across attributes; (C) Evolution of attribute-liking correlations from initial to post-interaction evaluations.}
    \label{fig:correlation}
    \vspace{-5pt}
\end{figure}

We evaluate our system's ability to model four key dimensions of social dynamics evolution: partner preference changes, self-perception adaptation, external evaluation shifts, and match decision accuracy. As illustrated in Figure~\ref{fig:correlation}, our analysis reveals several important patterns that characterize human social dynamics and demonstrates the remarkable similarity in evolution patterns between human participants and our GNWT-Agents. 

\textbf{For partner preferences (Figure~\ref{fig:correlation}A)}, both humans and agents show consistent patterns in attribute importance shifts. Attractiveness importance increases substantially ($+39.0\%$ human, $+25.0\%$ agent), while intelligence ($-24.8\%$ human, $-15.2\%$ agent) and sincerity ($-16.6\%$ human, $-10.5\%$ agent) show significant decreases. Shared interests show comparable positive changes ($+9.8\%$ human, $+9.7\%$ agent), while fun remains relatively stable with slight increases ($+1.3\%$ human, $+5.8\%$ agent). Ambition ($-7.0\%$ human, $-4.5\%$ agent) exhibits moderate decreases, suggesting its relative stability as a core value less influenced by short-term interactions.

\textbf{For self-perception (Figure~\ref{fig:correlation}B)}, there is subtle but consistent calibration of traits through social interaction. Unlike partner preferences, self-perception shows more conservative adjustments, with small negative shifts across most dimensions for both humans and agents: attractiveness ($+0.3\%$ human, $-0.5\%$ agent), sincerity ($-3.5\%$ human, $-2.5\%$ agent), intelligence ($-1.9\%$ human, $-1.2\%$ agent), fun ($-1.3\%$ human, $-0.8\%$ agent), and ambition ($-0.8\%$ human, $-0.5\%$ agent). The self-other perception gap narrows consistently for both humans and agents (from $0.8\rightarrow0.7$ human, $0.9\rightarrow0.7$ agent), reflecting the social calibration process through which external feedback helps align self-image with social reality.

\begin{figure}
    \centering
    \includegraphics[width=0.9\linewidth]{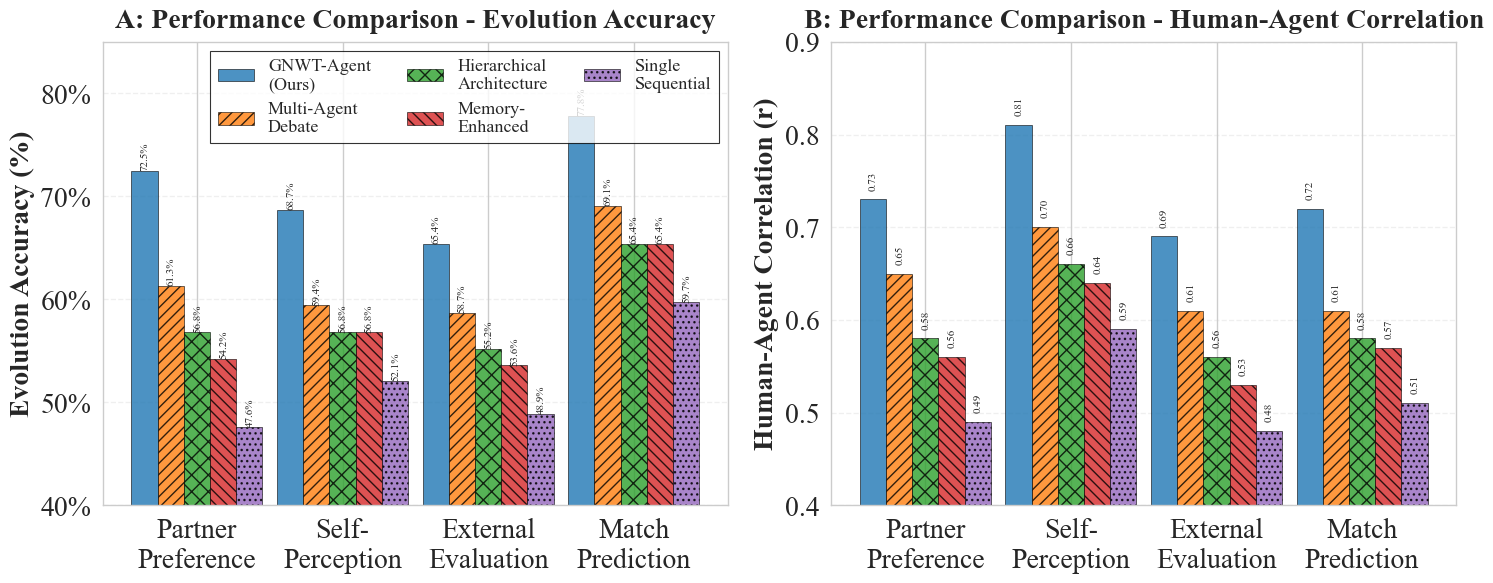}
    \caption{Comparison of GNWT-Agent with baseline methods across evolution dimensions. (A) Evolution accuracy showing GNWT-Agent's superior performance in all metrics; (B) Human-agent correlation demonstrating stronger alignment with human data compared to baseline approaches.}
    \label{fig:comparative}
    \vspace{-5pt}
\end{figure}

\textbf{For external evaluation} (Figure~\ref{fig:correlation}C), there is a dramatic shift in evaluation criteria from Time 1 to Time 2. For humans, the initially strong correlations between attributes and overall liking at Time 1 (ranging from $r=0.44$ to $r=0.69$) diminish dramatically at Time 2 (ranging from $r=-0.06$ to $r=0.07$), suggesting a fundamental change in evaluation criteria following interaction. Our GNWT-Agent shows a similar pattern, with high Time 1 correlations (ranging from $r=0.40$ to $r=0.65$) decreasing substantially at Time 2, though maintaining slight positive correlations (ranging from $r=0.08$ to $r=0.20$). This pattern indicates that both humans and agents undergo significant shifts in their evaluation frameworks through social interaction, though agents retain more of their initial criteria than humans do.

\textbf{Comparative Analysis:} As shown in Figure~\ref{fig:comparative}A, our GNWT-Agent consistently outperforms all baseline methods across evolution dimensions, with particularly strong advantages in partner preference evolution (72.5\% vs. 61.3\% for Multi-Agent Debate) and match prediction accuracy (77.8\% vs. 69.1\%). Figure~\ref{fig:comparative}B further demonstrates our system's superior human-agent correlation, with GNWT-Agent achieving strong correlation values above 0.7 in multiple dimensions, while baseline methods fall below this threshold. The complete comparative data tables with detailed metrics and standard deviations can be found in Appendix~\ref{subsec:detailed_results}.

\subsection{Human Validation Studies}
\label{subsec:human_validation}

To validate our approach, we conducted studies where real people evaluated AI versions of themselves ("digital twins") across dating and job interview contexts. Our interactive Adventure-Based Personality Assessment (Algorithm~\ref{alg:adventure_assessment}) infers personality traits through a series of immersive, scenario-driven choices. This approach mitigates the self-presentation bias often found in traditional self-report questionnaires by observing behavior rather than relying on self-description. The resulting trait estimates show strong correlation with established Big Five scores ($r = 0.82$), as detailed in Appendix~\ref{appendix:adventure_example}, and offer a more authentic reflection of participants’ personalities in real-world contexts.

We conducted a Speed Dating Study with 20 participants who watched recordings of their AI twin in simulated speed dates, and a Job Interview Study with 10 different participants who observed their twins respond to interview questions. Table~\ref{tab:human_verification} summarizes our findings. In the dating context, participants rated their twin's behavioral fidelity at 5.6 out of 7 (SD=0.8), significantly above the neutral point of 4 (p<0.001). For decision concordance, participants agreed with their twin's choices 74\% of the time when presented with the same scenarios.

The job interview study showed slightly better results, with behavioral fidelity rated at 5.8/7 (SD=0.6) and decision concordance at 81\%. Our system demonstrated appropriate contextual adjustments: in job interviews compared to dating scenarios, we observed more influence from the Planning component (31\% vs. 22\%), more consistent emotional processing (fluctuations of 3\% vs. 7\%), and greater adherence to professional norms during formal segments. These results demonstrate that our system creates recognizable digital representations that adapt appropriately to different social contexts while maintaining core personality characteristics, with somewhat stronger performance in professional settings.

\subsection{Limitations and Future Work}
\label{subsec:limitations}

While our system demonstrated strong performance, we identified several enhancement opportunities. Fine-tuning module calibration could improve agent fidelity, as minor imbalances between components affected prediction accuracy. Our architecture's cross-cultural performance could be enhanced through additional cultural parameters in the Social\_norms module. Future work should incorporate non-verbal communication patterns and optimize computational efficiency for larger populations. These enhancements require primarily configuration adjustments rather than redesign, enabling applications across professional collaboration, human-AI teams, and personalized education.

\section{Broader Impacts \& Conclusion}
\label{sec:broader_impacts}

Our GNWT-Agent architecture and CogniPair system offer significant positive societal implications. In dating applications, digital twins can efficiently identify compatible matches based on psychological alignment rather than superficial attributes. For hiring contexts, our approach helps candidates assess position fit while enabling companies to identify applicants whose cognitive styles would integrate well with existing teams. The architecture also enables enhanced human-AI collaboration in education, healthcare, and professional development, while contributing to cognitive science through computational modeling of psychological processes. While acknowledging the need for privacy safeguards and bias prevention measures, we believe this technology's positive potential substantially outweighs potential concerns. We presented the first computational implementation of Global Workspace Theory for AI agents, demonstrating psychological realism through specialized cognitive modules coordinated via a workspace broadcast mechanism. Our approach bridges psychological and social behavior gaps in current AI systems, with strong correlations to human behavior across multiple contexts. This work establishes a foundation for digital twins that maintain consistent personality while naturally evolving through social experience, enabling more authentic AI assistants across domains requiring genuine psychological processes.

\bibliographystyle{plain}
\bibliography{references}

\newpage
\appendix

\section{Appendix}
\label{sec:appendix}

\section*{Appendix Contents}
\vspace{0.5em}
\renewcommand{\descriptionlabel}[1]{\hspace{\labelsep}\textbf{#1}}
\begin{description}[leftmargin=0em, labelindent=0em]
    \item[A.1 Agent Architecture] \hfill \textit{Page~\pageref{subsec:agent_architecture}} \\
    \small{Internal structure and module flow.}
    
    \item[A.2 Algorithms and Pseudocode] \hfill \textit{Page~\pageref{subsec:algorithms}} \\
    \small{Key algorithmic procedures and decision logic.}
    
    \item[A.3 System Prompt Flow] \hfill \textit{Page~\pageref{subsec:system_prompt}} \\
    \small{Breakdown of prompt construction and token broadcast paths.}
    
    \item[A.4 Worked Example: Interview Anxiety] \hfill \textit{Page~\pageref{subsec:worked_example}} \\
    \small{End-to-end walk-through of a real-world query scenario.}
    
    \item[A.5 Use Case: Dating Application] \hfill \textit{Page~\pageref{subsec:usecase_dating}} \\
    \small{Evaluation results in interpersonal matching simulations.}
    
    \item[A.6 Use Case: Job Application] \hfill \textit{Page~\pageref{subsec:usecase_job}} \\
    \small{Adaptation of the framework for career placement tasks.}
    
    \item[A.7 Adventure-Based Personality Assessment] \hfill \textit{Page~\pageref{subsec:adventure_example}} \\
    \small{Roleplay-driven method to uncover latent traits.}
    
    \item[A.8 Detailed Results] \hfill \textit{Page~\pageref{subsec:detailed_results}} \\
    \small{Full quantitative results and evaluation tables.}
\end{description}
\vspace{0.5em}
\rule{\textwidth}{0.5pt}

\newpage

\newpage
\subsection{Agent Architecture}
\label{subsec:agent_architecture}

\begin{figure}[h]
    \centering
    \includegraphics[width=1\linewidth]{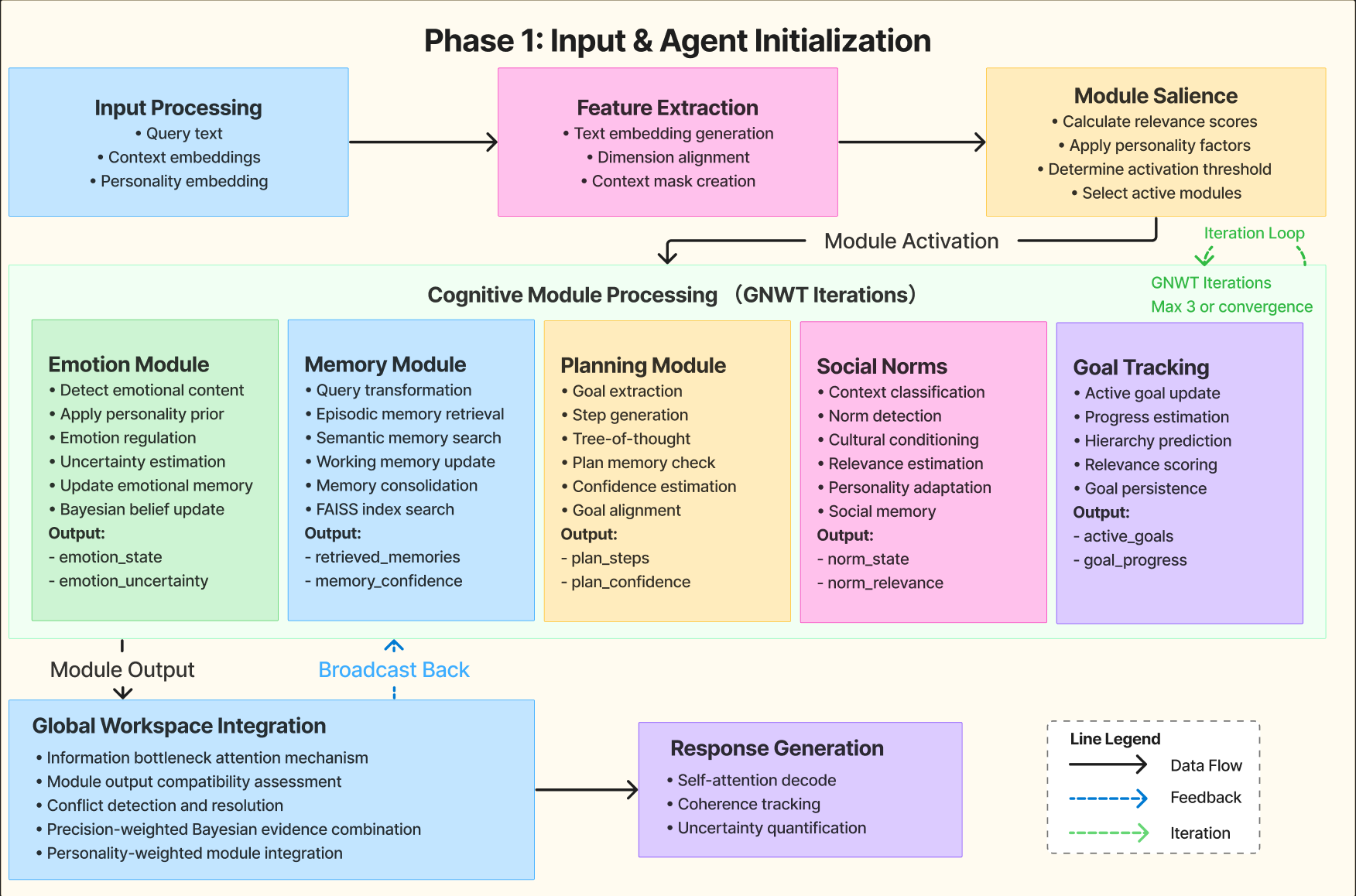}
    \caption{GNWT-Agent Architecture Internal Processing Flow}
    \label{fig:GNWT-Agent_architecture}
\end{figure}

\subsubsection{LLM Details}
\label{ssec:hyperparams}

We used the OpenAI API for GPT-4o with top\_p set to 1, max\_tokens set to 200, min\_tokens set to 0, and temperature set to 0.9 (with all other parameters at their default values),

\newpage
\subsection{Algorithms and Pseudocode}
\label{subsec:algorithms}

\begin{algorithm*}
\footnotesize
\caption{GNWT-Agent: GNWT-Based Social Pairing Agent}
\label{alg:cognipair}
\begin{algorithmic}[1]
\Require User query $Q$, conversation history $H$, agent personality profile $P$, maximum iterations $T$, convergence threshold $\epsilon$
\State $\text{Modules} \gets \{\text{Emotion}, \text{Memory}, \text{Planning}, \text{SocialNorms}, \text{GoalTracking}\}$
\State $\text{Modules} \gets \text{InitializeModules}(P)$ \Comment{Initialize modules based on personality traits}
\State $\mathrm{GW} \gets \text{InitializeGlobalWorkspace}(P, H)$
\State $\mathrm{SC} \gets \text{InitializeSalienceCalculator}(P)$
\State $R_{\text{prev}} \gets \emptyset$ \Comment{Store previous module responses}

\For{$t = 1 \to T$}
  \State \textbf{parallel for} each module $M \in \mathrm{Modules}$:
  \indent 
  \State $R_M \gets M.\textsc{process}(Q, H, \mathrm{GW})$ \Comment{Module's local response}

  \ForAll{$M \in \mathrm{Modules}$}
    \State $S_M \gets \mathrm{SC}.\textsc{evaluate}(R_M, P)$ \Comment{Compute salience score}
  \EndFor
  
  \State $M^* \gets \arg\max_{M} S_M$ \Comment{Select most salient module}
  \State $C \gets R_{M^*}$ \Comment{Content to broadcast}
  
  \If{$\exists M_i, M_j : \text{Conflict}(R_{M_i}, R_{M_j}) > \text{threshold}$}
    \State $C \gets \textsc{ResolveConflict}(R_{M_i}, R_{M_j}, \mathrm{GW}, P)$ \Comment{Resolve module conflicts}
  \EndIf
  
  \State $\textsc{ignite}(\mathrm{GW}, C)$ \Comment{Activate content in workspace}
  \State $\textsc{broadcast}(C)$ to all modules
  
  \ForAll{$M \in \mathrm{Modules}$}
    \State $M.\textsc{update}(C, \mathrm{GW})$ \Comment{Update module state}
  \EndFor
  
  \If{$\max_{M} |R_M - R_{\text{prev},M}| < \epsilon$ \textbf{or} $t = T$} \Comment{Minimal change in responses}
    \State \textbf{break}
  \EndIf
  
  \State $R_{\text{prev}} \gets \{R_M | M \in \text{Modules}\}$ \Comment{Save current responses for next iteration}
\EndFor

\State $\text{Response} \gets \textsc{IntegrateModuleOutputs}(\{R_M | M \in \text{Modules}\}, \mathrm{GW}, P)$
\State $\text{Memory}.\textsc{updateLongTerm}(Q, \text{Response}, H)$ \Comment{Update long-term memory}
\State $P \gets \textsc{UpdatePreferences}(P, Q, \text{Response}, H)$ \Comment{Adaptively update preferences}

\State \textbf{return} $\text{Response}$ \Comment{Return final integrated response}
\end{algorithmic}
\end{algorithm*}
\newpage



\begin{algorithm*}
\footnotesize
\caption{CogniPair: Speed Dating Cognitive Simulation System}
\label{alg:dating_simulation}
\begin{algorithmic}[1]
\Require Agent profiles $\mathcal{P} = \{P_1, P_2, \ldots, P_n\}$, batch size $b$, cognitive parameters $\Theta$
\Ensure Dating results $\mathcal{R}$, cognitive trace data $\mathcal{T}$, matches $\mathcal{M}$

\State $\mathcal{A} \gets \emptyset$ \Comment{Empty agent set}
\State $\mathcal{M} \gets \emptyset$ \Comment{Empty matches set}
\State $\mathcal{T} \gets \emptyset$ \Comment{Empty cognitive trace set}

\For{$P_i \in \mathcal{P}$} \Comment{Initialize all agents}
    \State $\mathrm{Modules}_i \gets \{\text{Emotion}, \text{Memory}, \text{Planning}, \text{SocialNorms}, \text{Attraction}\}$
    \State $\mathrm{GW}_i \gets \text{InitializeGlobalWorkspace}(P_i)$ \Comment{Initialize global workspace}
    \State $\mathrm{DA}_i \gets \text{InitializeDatingAttributes}(P_i)$ \Comment{Map dating preferences}
    \State $A_i \gets \text{CognitiveAgent}(P_i, \mathrm{Modules}_i, \mathrm{GW}_i, \mathrm{DA}_i, \Theta)$
    \State $\mathcal{A} \gets \mathcal{A} \cup \{A_i\}$
\EndFor

\State $\mathcal{P}_{pairs} \gets \text{GenerateCompatiblePairs}(\mathcal{A})$ \Comment{Based on gender/orientation}
\State $\mathcal{B} \gets \text{BatchPairs}(\mathcal{P}_{pairs}, b)$ \Comment{Create batches of size b}

\For{batch $B \in \mathcal{B}$}
    \State $\text{Results}_B \gets \emptyset$
    \For{pair $(A_i, A_j) \in B$} \textbf{parallel} \Comment{Process pairs in parallel}
        \State $H_{ij} \gets \emptyset$ \Comment{Empty conversation history}
        \State $C_{ij} \gets \text{InitializeContext}(A_i, A_j)$ \Comment{Setting, shared context}
        
        \For{$r = 1 \to \text{MAX\_ROUNDS}$} \Comment{Conversation rounds}
            \State $Q_i \gets A_i.\textsc{GenerateQuery}(H_{ij}, C_{ij})$ \Comment{Generate question/statement}
            \State $\mathcal{T} \gets \mathcal{T} \cup A_i.\mathrm{GW}.\textsc{getTrace}()$ \Comment{Capture cognitive trace}
            
            \State $H_{ij} \gets H_{ij} \cup \{(i, Q_i)\}$ \Comment{Update conversation history}
            
            \State $R_j \gets A_j.\textsc{GenerateResponse}(Q_i, H_{ij}, C_{ij})$ \Comment{Using Alg.~\ref{alg:cognipair}}
            \State $\mathcal{T} \gets \mathcal{T} \cup A_j.\mathrm{GW}.\textsc{getTrace}()$ \Comment{Capture cognitive trace}
            
            \State $H_{ij} \gets H_{ij} \cup \{(j, R_j)\}$ \Comment{Update conversation history}
            
            \If{$r < \text{MAX\_ROUNDS}$}
                \State $Q_j \gets A_j.\textsc{GenerateQuery}(H_{ij}, C_{ij})$
                \State $\mathcal{T} \gets \mathcal{T} \cup A_j.\mathrm{GW}.\textsc{getTrace}()$
                
                \State $H_{ij} \gets H_{ij} \cup \{(j, Q_j)\}$
                
                \State $R_i \gets A_i.\textsc{GenerateResponse}(Q_j, H_{ij}, C_{ij})$ \Comment{Using Alg.~\ref{alg:cognipair}}
                \State $\mathcal{T} \gets \mathcal{T} \cup A_i.\mathrm{GW}.\textsc{getTrace}()$
                
                \State $H_{ij} \gets H_{ij} \cup \{(i, R_i)\}$
            \EndIf
            
            \State $A_i.\textsc{UpdateAttraction}(H_{ij}, A_j)$ \Comment{Update attraction dynamics}
            \State $A_j.\textsc{UpdateAttraction}(H_{ij}, A_i)$
        \EndFor
        
        \State $S_i \gets A_i.\textsc{EvaluateCompatibility}(A_j, H_{ij})$ \Comment{Final decision}
        \State $S_j \gets A_j.\textsc{EvaluateCompatibility}(A_i, H_{ij})$
        
        \State $\text{Results}_B \gets \text{Results}_B \cup \{(A_i, A_j, S_i, S_j, H_{ij})\}$
        
        \If{$S_i \geq \text{THRESHOLD} \land S_j \geq \text{THRESHOLD}$}
            \State $\mathcal{M} \gets \mathcal{M} \cup \{(A_i, A_j)\}$ \Comment{Record match}
        \EndIf
        
        \State $A_i.\textsc{UpdatePreferences}(A_j, H_{ij}, S_i, S_j)$ \Comment{Preference evolution}
        \State $A_j.\textsc{UpdatePreferences}(A_i, H_{ij}, S_j, S_i)$
    \EndFor
    \State $\mathcal{R} \gets \mathcal{R} \cup \text{Results}_B$
\EndFor

\State \textbf{return} $(\mathcal{R}, \mathcal{T}, \mathcal{M})$
\end{algorithmic}
\end{algorithm*}

\newpage


\begin{algorithm}
\caption{Adventure-Based Personality Assessment}
\label{alg:adventure_assessment}
\begin{algorithmic}[1]
\State Initialize \texttt{personalityProfile} and \texttt{traitConfidence}
\State \texttt{modelPreference} = GetUserModelPreference() \Comment{Cloud (GPT-4o) or Local (Ollama)}

\While{more scenarios needed AND trait coverage insufficient}
    \State Select and present next scenario from pool
    \State Collect user's choice
    
    \If{\texttt{modelPreference} == "local"}
        \State Analyze choice using Ollama (llama3 or deepseek-r1)
    \Else
        \State Analyze choice using GPT-4o
    \EndIf
    
    \State Update personality traits based on LLM analysis
    
    \State Generate follow-up question based on user's choice
    \State Collect user's free-text response
    
    \If{\texttt{modelPreference} == "local"}
        \State Analyze free text using Ollama
    \Else
        \State Analyze free text using GPT-4o
    \EndIf
    
    \State Update personality traits based on text analysis
\EndWhile

\State Normalize and validate final personality profile
\State \Return \texttt{finalProfile}
\end{algorithmic}
\end{algorithm}
\newpage

\newpage
\subsection{Detailed System Flow Analysis: GNWT-Agent Cognitive Architecture}
\label{subsec:system_prompt}

\label{appendix:gnwt_agent}

This appendix presents a step-by-step analysis of information flow through the GNWT-Agent cognitive architecture. GNWT-Agent's central innovation is its hybrid neural-symbolic approach that combines specialized neural modules with LLM reasoning via a global workspace mechanism, implementing a neurobiologically-informed cognitive architecture.

\subsection{Formal Architecture Definition}

The GNWT-Agent cognitive architecture is formally defined as a quintuple:

\begin{equation}
\mathcal{E} = (\mathcal{M}, \mathcal{W}, \mathcal{I}, \mathcal{L}, \mathcal{P})
\label{eq:architecture_definition}
\end{equation}

Where:
\begin{description}
    \item[$\mathcal{M}$] $= \{M_{\text{emo}}, M_{\text{mem}}, M_{\text{plan}}, M_{\text{norm}}, M_{\text{goal}}\}$ represents the set of specialized cognitive modules
    \item[$\mathcal{W}$] denotes the global workspace integration mechanism
    \item[$\mathcal{I}$] signifies the information bottleneck attention system
    \item[$\mathcal{L}$] represents the language model interface
    \item[$\mathcal{P}$] characterizes the personality representation space
\end{description}

Each cognitive module $M_i \in \mathcal{M}$ implements a hybrid neural-symbolic architecture:

\begin{equation}
M_i = (\mathcal{N}_i, \mathcal{L}_i, \mathcal{T}_i, \mathcal{D}_i, \mathcal{I}_i, \mathcal{S}_i)
\label{eq:module_definition}
\end{equation}

Where:
\begin{description}
    \item[$\mathcal{N}_i$] denotes the neural processing component
    \item[$\mathcal{L}_i$] signifies the module-specific LLM component
    \item[$\mathcal{T}_i$] represents the tensor-text conversion mechanism
    \item[$\mathcal{D}_i$] denotes the differentiable memory system
    \item[$\mathcal{I}_i$] signifies the module interface specification
    \item[$\mathcal{S}_i$] represents the salience computation function
\end{description}

\begin{table}[htbp]
\centering
\caption{Specialized Modules and Their Neuroanatomical Bases}
\label{table:modules}
\begin{tabular}{@{}lll@{}}
\toprule
\textbf{Module} & \textbf{Function} & \textbf{Neural Inspiration} \\
\midrule
Emotion ($M_{\text{emo}}$) & Affective processing & Limbic system, amygdala, insula \\
Memory ($M_{\text{mem}}$) & Information retrieval & Hippocampus, temporal cortex \\
Planning ($M_{\text{plan}}$) & Structured reasoning & Frontopolar cortex, DLPFC \\
Social Norms ($M_{\text{norm}}$) & Social context & mPFC, TPJ \\
Goal Tracking ($M_{\text{goal}}$) & Hierarchical goals & OFC, ACC \\
\bottomrule
\end{tabular}
\end{table}

\subsection{Information Flow Process}

\subsubsection{Initial Text Encoding}

The information flow begins with the transformation of text inputs into neural representations:

\begin{equation}
\begin{aligned}
e_Q &= \phi_{\text{embed}}(Q) \in \mathbb{R}^d \\
e_H &= \{\phi_{\text{embed}}(h_i) | h_i \in H\} \in \mathbb{R}^{n \times d} \\
e_P &= \psi(p) \in \mathbb{R}^{d_p}
\end{aligned}
\label{eq:text_encoding}
\end{equation}

Where $\phi_{\text{embed}}$ is the embedding model that converts text to dense vectors, $Q$ is the query text, $H$ is the conversation history, and $p$ is the personality profile.

\subsubsection{Feature Extraction and Embedding Alignment}

The raw embeddings undergo feature extraction and alignment:

\begin{equation}
\begin{aligned}
e'_Q &= \text{FeatureExtractor}(e_Q) \\
e'_H &= \{\text{FeatureExtractor}(e_{h}) | e_{h} \in e_H\}
\end{aligned}
\label{eq:feature_extraction}
\end{equation}

\subsubsection{Module Salience Calculation}

Each module calculates its relevance to the current input:

\begin{equation}
\begin{aligned}
s_i &= \mathcal{S}_i(e'_Q, e'_H, e_P, \mathcal{G}) \\
&= \alpha_i + \sum_j \beta_{ij} \cdot f_{ij}(e'_Q, e'_H, e_P, \mathcal{G})
\end{aligned}
\label{eq:salience}
\end{equation}

Where $\alpha_i$ is the baseline salience, $\beta_{ij}$ are weighting coefficients, $f_{ij}$ are feature extractors, and $\mathcal{G}$ is the current global workspace state.

\subsubsection{Parallel Module Processing}
\label{subsec:parallel_processing}

Each module independently processes the input using a hybrid neural-LLM approach with the following general pattern:

\begin{enumerate}
    \item \textbf{Neural Processing}: Extract relevant features and apply module-specific transformations
    \item \textbf{Tensor-to-Text Conversion}: Convert neural representations to LLM-readable format
    \item \textbf{LLM Processing}: Generate structured symbolic representations using prompting
    \item \textbf{Text-to-Tensor Conversion}: Transform LLM outputs back to neural representations
    \item \textbf{Output Integration}: Combine neural and symbolic components for module output
\end{enumerate}

Module-specific processing includes:

\begin{itemize}
    \item \textbf{Emotion Module}: Detects emotional states and regulation strategies
    \item \textbf{Memory Module}: Retrieves and integrates episodic, semantic, and working memories
    \item \textbf{Planning Module}: Generates structured plans for achieving identified goals
    \item \textbf{Social Norms Module}: Identifies appropriate social contexts and behavioral norms
    \item \textbf{Goal Tracking Module}: Maintains hierarchical goal representations and tracks progress
\end{itemize}

\subsubsection{Global Workspace Integration}
\label{subsec:global_integration}

Module outputs are projected to a common workspace dimension and integrated based on salience:

\begin{algorithm}[ht]
\caption{Global Workspace Integration}
\label{alg:workspace_integration}
\begin{algorithmic}[1]
\State $\text{module\_rep}_i \gets \text{ProjectToWorkspace}(\text{output}_i)$ for each module $i$
\State $\text{normalized\_salience} \gets \text{Softmax}(s)$
\State $\text{personality\_weights} \gets \text{Softmax}(\text{PersonalityToIntegration}(e_P))$
\State $\text{combined\_weights} \gets \frac{\text{normalized\_salience} + \text{personality\_weights}}{2}$

\For{each pair of modules $(i,j)$ where $i \neq j$}
    \State $\text{conflict\_scores}_{i,j} \gets \text{ConflictDetector}([\text{module\_rep}_i, \text{module\_rep}_j])$
\EndFor

\State Apply conflict resolution adjustments to weights
\State $\text{integrated\_output} \gets \sum_i \text{adjusted\_weights}_i \cdot \text{module\_rep}_i$

\If{major conflicts detected}
    \State Apply specialized conflict resolution
\EndIf

\State $\text{workspace\_state} \gets \text{WorkspaceProjection}(\text{integrated\_output})$
\State Perform integration LLM call to further refine integration
\end{algorithmic}
\end{algorithm}

\subsubsection{Final Prompt Construction}

The global workspace state informs the construction of the final prompt for the response LLM, transforming module outputs into a coherent instruction format:

\begin{equation}
\begin{aligned}
\text{P}_{\text{system}} = \text{JoinWithNewlines}(&\text{PersonalityPrompt}(e_P),\\
&\text{ModulePrompts}(\{\text{output}_i\}, \{\text{adjusted\_weights}_i\}),\\
&\text{StrategyPrompt}(\text{integrated\_state}),\\
&\text{ConflictPrompt}(\text{conflicts\_resolved}),\\
&\text{ResponsePlanPrompt}(\text{response\_plan}))
\end{aligned}
\label{eq:system_prompt}
\end{equation}

\subsubsection{Response Generation and Memory Update}

The final response is generated and memory systems are updated:

\begin{equation}
\begin{aligned}
\text{history} &= [\text{prev\_messages}] \\
\text{query\_message} &= \{\text{"role": "user", "content": } Q\} \\
\text{full\_history} &= \text{history } + [\text{query\_message}] \\
R &= \text{LLM}_{\text{response}}(P_{\text{system}}, \text{full\_history}, \text{temperature}=0.7)
\end{aligned}
\label{eq:response_generation}
\end{equation}

\begin{equation}
\begin{aligned}
\text{EpisodicMemory.write}(&\text{Embed}(Q, R), \text{StructuredEncoding}(Q, R, \text{workspace\_state})) \\
\text{WorkingMemory.update}(&Q, R)
\end{aligned}
\label{eq:memory_update}
\end{equation}

\subsection{Processing Example}

The following example demonstrates the architecture's response to: "I'm really nervous about my job interview tomorrow. I've been preparing for weeks, but I still don't feel ready. What should I do?"

\begin{tcolorbox}[title=Module Salience Calculation, colback=yellow!5!white, colframe=yellow!75!black]
\small
\textbf{Salience Scores:}
\begin{itemize}
\item Emotion Module: 0.75 (anxiety markers, moderate neuroticism, emotional shift)
\item Planning Module: 0.67 (clear goal, structured task, high conscientiousness)
\item Memory Module: 0.52 (relevant experiences, moderate openness)
\item Social Norms Module: 0.48 (professional context, formal situation)
\item Goal Tracking Module: 0.45 (career advancement, interview milestone)
\end{itemize}

\textbf{Normalized weights:} [0.31, 0.28, 0.21, 0.12, 0.08]
\end{tcolorbox}

\begin{tcolorbox}[title=Global Workspace Integration, colback=gray!5!white, colframe=gray!75!black]
\small
\textbf{Conflict Detection:}
\begin{itemize}
\item Primary conflict: Emotion module (anxiety validation) vs. Planning module (action steps)
\end{itemize}

\textbf{Adjusted Weights:} [0.29, 0.25, 0.22, 0.14, 0.10]

\textbf{Integration Results:}
\begin{itemize}
\item Primary response focus: balanced emotional and practical support
\item Emotional context: anxiety (valence: -0.48, arousal: 0.68) with determination
\item Response strategy: validate anxiety then provide structured help
\item Conflict resolution: Balance validation with actionable steps
\end{itemize}

\textbf{Response Plan:}
\begin{enumerate}
\item Validate interview anxiety as normal
\item Acknowledge preparation effort positively
\item Provide targeted high-impact preparation steps
\item Include confidence-building perspective
\item End with encouragement and well-wishes
\end{enumerate}
\end{tcolorbox}

\begin{tcolorbox}[title=Final Response, colback=blue!5!white, colframe=blue!75!black]
\small
It's completely normal to feel nervous before an interview, especially for a company
you're really excited about! The fact that you've been preparing for weeks shows your
dedication, and that preparation won't disappear overnight, even if you're feeling
anxious right now.

At this point, here's what I'd suggest focusing on:

1. Review your strongest points - skim through the key qualifications and 2-3 
   accomplishment stories you want to highlight. Don't try to memorize everything.

2. Practice answering 3-5 common questions out loud - this builds muscle memory that
   helps even when nerves kick in.

3. Prepare your logistics for tomorrow - lay out your outfit, plan your route, and
   gather any materials you need to bring.

4. Most importantly, make sure you get adequate rest tonight. A clear mind will serve
   you better than a few more hours of anxious preparation.

Remember that interviews are a two-way conversation, not an examination. You've been
preparing for weeks - now it's time to trust that preparation and focus on presenting
your authentic self.

Good luck tomorrow! You've put in the work, and that's the most important part.
\end{tcolorbox}

\subsection{Module LLM Call Architecture}

The GNWT-Agent architecture employs a cascade of LLM calls with neural pre-processing and post-processing:

\begin{equation}
\begin{aligned}
\text{Module-Level LLM Calls}: \{
&\text{LLM}_{\text{emo}}(P_{\text{emo}}), 
\text{LLM}_{\text{mem}}(P_{\text{mem}}), 
\text{LLM}_{\text{plan}}(P_{\text{plan}}),\\
&\text{LLM}_{\text{norm}}(P_{\text{norm}}), 
\text{LLM}_{\text{goal}}(P_{\text{goal}})
\}
\end{aligned}
\label{eq:module_llm_calls}
\end{equation}

\begin{equation}
\text{Integration LLM Call}: \text{LLM}_{\text{integration}}(P_{\text{integration}})
\label{eq:integration_llm_call}
\end{equation}

\begin{equation}
\text{Response LLM Call}: \text{LLM}_{\text{response}}(P_{\text{system}}, Q)
\label{eq:response_llm_call}
\end{equation}

This creates a complete cognitive cycle:
\begin{equation}
\text{Input} \rightarrow \text{Neural Processing} \rightarrow \text{Module LLMs} \rightarrow \text{Integration LLM} \rightarrow \text{Response LLM} \rightarrow \text{Memory Update}
\label{eq:cognitive_cycle}
\end{equation}

\subsection{Summary}

The GNWT-Agent cognitive architecture implements a hybrid neural-symbolic approach that integrates multiple specialized modules within a neurobiologically-informed framework. The architecture's core components include five specialized cognitive modules addressing emotion, memory, planning, social norms, and goals.

The information flow through the system follows a comprehensive process:
\begin{enumerate}
    \item Initial text encoding via dense vector representations
    \item Parallel processing across specialized modules
    \item Dynamic salience-based resource allocation
    \item Global workspace integration with conflict detection and resolution
    \item Structured prompt construction
    \item Final response generation
    \item Memory update and continuous learning
\end{enumerate}

\begin{table}[ht]
\centering
\caption{Strengths and Limitations of GNWT-Agent Architecture}
\label{table:strengths_limitations}
\begin{tabular}{@{}p{0.48\textwidth}p{0.48\textwidth}@{}}
\toprule
\textbf{Strengths} & \textbf{Limitations} \\
\midrule
Enhanced interpretability through explicit module contributions & Computational complexity due to multiple LLM calls \\
Improved uncertainty handling via Bayesian uncertainty propagation & Challenges in consistent knowledge representation \\
Consistent personality representation through dedicated space & Need for further empirical validation \\
Dynamic adaptation through salience-based processing & Potential for response latency issues \\
Neurobiologically-inspired cognitive processing & Complex architecture requiring extensive fine-tuning \\
\bottomrule
\end{tabular}
\end{table}

In conclusion, the GNWT-Agent cognitive architecture represents a significant step toward creating AI systems that not only process language effectively but do so through mechanisms that more closely approximate human cognitive processes, potentially leading to more natural, adaptive, and comprehensible AI interactions.

\newpage
\subsection{Worked Example: Interview Anxiety Query}
\label{subsec:worked_example}

This section presents a detailed step-by-step walkthrough of the complete GNWT-Agent cognitive architecture processing flow for a single example query. By tracing the transformations from raw input to final response, we provide a concrete illustration of the theoretical architecture described in previous sections.

\subsection{Example Input Query}

The example query represents a common scenario of pre-interview anxiety:

\begin{tcolorbox}[colback=gray!5, colframe=gray!40, title=User Input]
``I'm really nervous about my job interview tomorrow. I've been preparing for weeks, but I still don't feel ready. What should I do?''
\end{tcolorbox}

\subsection{Initial Text Encoding and Embedding}

The first transformation converts the raw text into numerical representations:

\begin{align}
\text{Input} &\xrightarrow{\phi_{\text{embed}}} \text{Query Embedding} \\
\text{``I'm really nervous...''} &\rightarrow [0.086, -0.143, 0.257, \ldots, 0.112] \in \mathbb{R}^{768}
\end{align}

\noindent
Previous conversation context is also encoded:
\begin{align}
\text{``I've been job hunting for months''} &\rightarrow [0.141, 0.092, \ldots] \in \mathbb{R}^{768} \\
\text{``I finally got an interview at my dream company''} &\rightarrow [0.235, -0.124, \ldots] \in \mathbb{R}^{768}
\end{align}

\noindent
Feature extraction enhances these raw embeddings:
\begin{align}
\text{Enhanced query embedding: } [0.127, -0.086, 0.313, \ldots] \in \mathbb{R}^{768}
\end{align}

\subsection{Module Salience Calculation}

Each module calculates its relevance to the query through multi-factor salience functions:

\begin{table}[h]
\centering
\begin{tabular}{lcccc}
\hline
\textbf{Module} & \textbf{Primary Factors} & \textbf{Values} & \textbf{Raw Salience} & \textbf{Normalized} \\
\hline
Emotion & \begin{tabular}{l}Emotional intensity \\ Personality factor \\ Emotional change\end{tabular} & \begin{tabular}{l}0.83 \\ 0.52 \\ 0.60\end{tabular} & 0.75 & 0.31 \\
\hline
Planning & \begin{tabular}{l}Goal clarity \\ Task structure \\ Personality factor\end{tabular} & \begin{tabular}{l}0.78 \\ 0.82 \\ 0.76\end{tabular} & 0.67 & 0.28 \\
\hline
Memory & \begin{tabular}{l}Memory match \\ Personality factor \\ Recency boost\end{tabular} & \begin{tabular}{l}0.64 \\ 0.61 \\ 0.35\end{tabular} & 0.52 & 0.21 \\
\hline
Social Norms & \begin{tabular}{l}Norm relevance \\ Formality \\ Personality factor\end{tabular} & \begin{tabular}{l}0.71 \\ 0.70 \\ 0.63\end{tabular} & 0.48 & 0.12 \\
\hline
Goal Tracking & \begin{tabular}{l}Goal relevance \\ Personality factor \\ Progress factor\end{tabular} & \begin{tabular}{l}0.76 \\ 0.65 \\ 0.58\end{tabular} & 0.45 & 0.08 \\
\hline
\end{tabular}
\caption{Module salience calculation for the interview anxiety query}
\label{tab:salience}
\end{table}

The emotional content of the query results in the Emotion Module having the highest salience (0.31), followed closely by the Planning Module (0.28).

\subsection{Parallel Module Processing}

\subsubsection{Emotion Module Processing}

The Emotion Module performs neural processing followed by LLM reasoning:

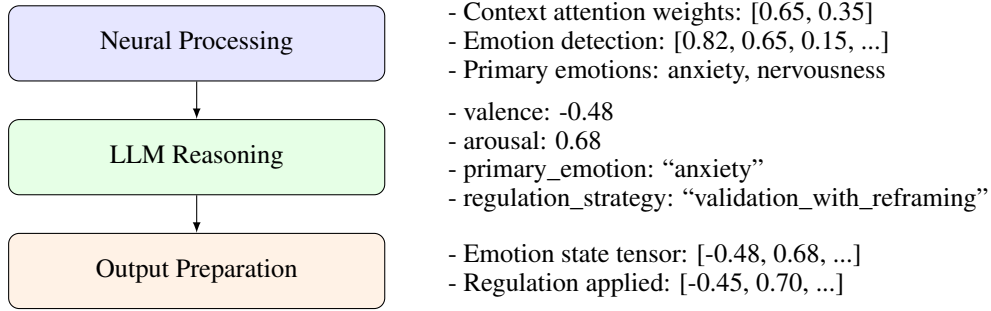
\begin{figure}[h]
\centering
\begin{tikzpicture}
\node[draw, rounded corners, fill=blue!10, minimum width=5cm, minimum height=1cm] (neural) {Neural Processing};
\node[draw, rounded corners, fill=green!10, minimum width=5cm, minimum height=1cm, below=0.5cm of neural] (llm) {LLM Reasoning};
\node[draw, rounded corners, fill=orange!10, minimum width=5cm, minimum height=1cm, below=0.5cm of llm] (output) {Output Preparation};

\draw[-latex] (neural) -- (llm);
\draw[-latex] (llm) -- (output);

\node[anchor=west, align=left, right=0.5cm of neural] {
\begin{tabular}{l}
- Context attention weights: [0.65, 0.35] \\
- Emotion detection: [0.82, 0.65, 0.15, ...] \\
- Primary emotions: anxiety, nervousness
\end{tabular}
};

\node[anchor=west, align=left, right=0.5cm of llm] {
\begin{tabular}{l}
- valence: -0.48 \\
- arousal: 0.68 \\
- primary\_emotion: ``anxiety'' \\
- regulation\_strategy: ``validation\_with\_reframing''
\end{tabular}
};

\node[anchor=west, align=left, right=0.5cm of output] {
\begin{tabular}{l}
- Emotion state tensor: [-0.48, 0.68, ...] \\
- Regulation applied: [-0.45, 0.70, ...]
\end{tabular}
};
\end{tikzpicture}
\caption{Emotion Module processing flow}
\label{fig:emotion}
\end{figure}

\subsubsection{Planning Module Processing}

The Planning Module extracts the goal and generates structured steps:

\begin{tcolorbox}[colback=green!5, colframe=green!40, title=Planning Module LLM Output]
\begin{verbatim}
{
  "requires_planning": true,
  "planning_type": "preparation_strategy",
  "steps": [
    {
      "content": "Review core interview preparation materials", 
      "confidence": 0.93
    },
    {
      "content": "Practice answering common questions aloud", 
      "confidence": 0.89
    },
    {
      "content": "Prepare concise examples of achievements", 
      "confidence": 0.86
    },
    {
      "content": "Implement anxiety reduction techniques", 
      "confidence": 0.82
    },
    {
      "content": "Plan interview logistics (route, outfit, materials)", 
      "confidence": 0.91
    },
    {
      "content": "Get adequate rest before interview", 
      "confidence": 0.85
    }
  ],
  "plan_feasibility": 0.88,
  "goal_alignment": 0.84,
  "plan_uncertainty": 0.16
}
\end{verbatim}
\end{tcolorbox}

\subsubsection{Memory Module Processing}

The Memory Module retrieves relevant episodic and semantic memories:

\begin{tcolorbox}[colback=blue!5, colframe=blue!40, title=Memory Module Content]
\begin{verbatim}
{
  "episodic_memories": [
    {"content": "User mentioned job hunting for months", 
    "confidence": 0.88},
    {"content": "User expressed excitement about interview at 
    dream company", 
     "confidence": 0.92},
    {"content": "User indicated spending significant time preparing", 
     "confidence": 0.85}
  ],
  "semantic_knowledge": [
    {"content": "Interview anxiety is normal and expected", 
    "confidence": 0.94},
    {"content": "Over-preparation can sometimes increase anxiety", 
     "confidence": 0.82},
    {"content": "Sleep quality affects interview performance", 
    "confidence": 0.89}
  ]
}
\end{verbatim}
\end{tcolorbox}

\subsubsection{Social Norms Module Processing}

The Social Norms Module identifies the appropriate context and behavioral expectations:

\begin{tcolorbox}[colback=purple!5, colframe=purple!40, title=Social Norms Module Content]
\begin{verbatim}
{
  "social_context": "professional_development",
  "appropriate_norms": [
    "acknowledge interview anxiety as normal",
    "validate preparation efforts",
    "provide practical confidence-building advice",
    "offer both emotional and practical support",
    "maintain appropriate emotional engagement"
  ],
  "formality_level": 0.65,
  "appropriate_tone": "supportive_professional"
}
\end{verbatim}
\end{tcolorbox}

\subsubsection{Goal Tracking Module Processing}

The Goal Tracking Module identifies and relates current goals to existing goals:

\begin{tcolorbox}[colback=yellow!5, colframe=yellow!40, title=Goal Tracking Module Content]
\begin{verbatim}
{
  "identified_goals": [
    {"content": "Successfully complete job interview", 
    "explicitness": 0.92},
    {"content": "Reduce pre-interview anxiety", "explicitness": 0.85}
  ],
  "existing_goals_relevance": [
    {"goal": "Secure desired employment", "relevance": 0.94},
    {"goal": "Develop career in chosen field", "relevance": 0.86}
  ]
}
\end{verbatim}
\end{tcolorbox}

\subsection{Global Workspace Integration}

The outputs from all five modules are integrated through the global workspace mechanism:

\subsubsection{Conflict Detection and Resolution}

The system identifies and resolves potential conflicts between module outputs:

\begin{table}[h]
\centering
\begin{tabular}{c|ccccc}
 & Emotion & Planning & Memory & Social & Goal \\
\hline
Emotion & 0.00 & \textbf{0.35} & 0.10 & 0.05 & 0.10 \\
Planning & \textbf{0.35} & 0.00 & 0.15 & 0.10 & 0.05 \\
Memory & 0.10 & 0.15 & 0.00 & 0.05 & 0.10 \\
Social & 0.05 & 0.10 & 0.05 & 0.00 & 0.05 \\
Goal & 0.10 & 0.05 & 0.10 & 0.05 & 0.00 \\
\end{tabular}
\caption{Conflict detection matrix showing highest conflict between Emotion and Planning modules}
\label{tab:conflict}
\end{table}

The primary conflict (0.35) exists between the Emotion Module's emphasis on validation and the Planning Module's focus on action steps. This conflict affects the modules' weights:

\begin{align}
\text{Original weights: } &[0.31, 0.28, 0.21, 0.12, 0.08] \\
\text{Conflict-adjusted weights: } &[0.29, 0.25, 0.22, 0.14, 0.10]
\end{align}

\subsubsection{Integration Result}

The Integration LLM synthesizes information from all modules into a coherent state:

\begin{tcolorbox}[colback=gray!5, colframe=gray!40, title=Integration LLM Output (Abbreviated)]
\begin{verbatim}
{
  "integrated_state": {
    "primary_response_focus": "balanced_emotional_practical",
    "emotional_context": {
      "emotion": "anxiety",
      "secondary": "determination",
      "valence": -0.48,
      "arousal": 0.68
    },
    "response_strategy": {
      "approach": "validate_anxiety_then_provide_structured_help",
      "tone": "supportive_professional",
      "formality": "moderate"
    },
    "response_plan": [
      "validate interview anxiety as normal",
      "acknowledge preparation effort positively",
      "provide targeted high-impact preparation steps",
      "include confidence-building perspective",
      "end with encouragement and well-wishes"
    ]
  },
  "conflicts_resolved": [
    {
      "conflict": "Emotion module emphasizes validation vs 
      Planning module focuses on action steps",
      "resolution": "Balance emotional validation with 
      practical action steps by
                    first acknowledging feelings then providing concrete, 
                    manageable next steps",
      "confidence": 0.88
    }
  ]
}
\end{verbatim}
\end{tcolorbox}

\subsection{Final Prompt Construction}

The integrated state is transformed into a structured system prompt for the response generation LLM:

\begin{tcolorbox}[colback=blue!5, colframe=blue!40, title=System Prompt]
\begin{verbatim}
You are a cognitive agent with the following personality:
MBTI Type: ENFJ
Big 5 Traits:
- Openness: high (0.70)
- Conscientiousness: high (0.80)
- Extraversion: moderate (0.50)
- Agreeableness: very high (0.90)
- Neuroticism: moderate-low (0.40)

[EMOTION MODULE] (Contribution: 0.29): Pay careful attention to the
anxiety and nervousness the user is experiencing about their upcoming
job interview. The emotions detected include high anxiety (0.82) with
a negative valence (-0.48) but also determination (0.55). 
Emotional content identified: "really nervous", "don't feel ready". 
Use a validation_with_reframing approach that acknowledges 
the anxiety while gently building confidence.

[PLANNING MODULE] (Contribution: 0.25): 
Provide structured, preparation_strategy advice for the user's 
interview preparation. Focus on high-impact preparation activities 
that can be completed in limited time. Consider including: reviewing 
core materials, practicing answers aloud, preparing achievement examples, 
reducing anxiety, handling logistics, and getting rest.

[MEMORY MODULE] (Contribution: 0.22): 
Reference these key points from conversation history: 
User has been job hunting for months,  this is an interview at their
"dream company," and they have already spent significant
time preparing. Also incorporate knowledge that interview 
anxiety is normaland that over-preparation can sometimes 
increase anxiety rather than helping.

[SOCIAL NORMS MODULE] (Contribution: 0.14): 
Maintain a supportive_professional
tone appropriate for career_development context. 
Acknowledge that interview
anxiety is normal while validating preparation efforts. 
Observe these norms:
validate emotions, show empathy, offer practical support, 
maintain appropriate emotional engagement.

Follow this response strategy: 
validate_anxiety_then_provide_structured_help with
a supportive_professional tone at moderate formality level.

When addressing conflicts: 
Balance emotional validation with practical action steps
by first acknowledging feelings then providing concrete, 
manageable next steps.

Process the query by considering each 
cognitive dimension above according to
its contribution weight. Follow this response plan:
1. Validate interview anxiety as normal
2. Acknowledge preparation effort positively
3. Provide targeted high-impact preparation steps
4. Include confidence-building perspective
5. End with encouragement and well-wishes
\end{verbatim}
\end{tcolorbox}

\subsection{Response Generation}

The Response LLM generates the final output based on the system prompt:

\begin{tcolorbox}[colback=green!5, colframe=green!40, title=Generated Response]
It's completely normal to feel nervous before an interview, especially for a company
you're really excited about! The fact that you've been preparing for weeks shows your
dedication, and that preparation won't disappear overnight, even if you're feeling
anxious right now.

At this point, here's what I'd suggest focusing on:

1. Review your strongest points - skim through the key qualifications and 2-3 
   accomplishment stories you want to highlight. Don't try to memorize everything.

2. Practice answering 3-5 common questions out loud - this builds muscle memory that
   helps even when nerves kick in.

3. Prepare your logistics for tomorrow - lay out your outfit, plan your route, and
   gather any materials you need to bring.

4. Most importantly, make sure you get adequate rest tonight. A clear mind will serve
   you better than a few more hours of anxious preparation.

Remember that interviews are a two-way conversation, not an examination. You've been
preparing for weeks - now it's time to trust that preparation and focus on presenting
your authentic self.

Good luck tomorrow! You've put in the work, and that's the most important part.
\end{tcolorbox}

\subsection{Information Flow Analysis}

Examining the example reveals several key aspects of the GNWT-Agent architecture:

\begin{enumerate}
    \item \textbf{Modular Specialization}: Each module contributes distinct information:
    \begin{itemize}
        \item Emotion Module provides emotional understanding and regulation strategy
        \item Planning Module offers concrete action steps
        \item Memory Module contributes contextual information and relevant knowledge
        \item Social Norms Module guides appropriate tone and interaction norms
        \item Goal Tracking Module ensures alignment with long-term objectives
    \end{itemize}
    
    \item \textbf{Salience-Based Processing}: The architecture dynamically allocates attention based on query characteristics, with emotional content receiving the highest weight in this anxiety-focused query.
    
    \item \textbf{Conflict Resolution}: The system explicitly identifies and resolves the tension between emotional validation and practical advice through a balanced approach.
    
    \item \textbf{Structured Response Planning}: The final response follows the five-step plan specified in the integration phase, creating a cohesive structure that addresses multiple dimensions of the query.
    
    \item \textbf{Personality Influence}: The system's responses reflect the specified personality characteristics, particularly high agreeableness (0.90) through the empathetic tone.
\end{enumerate}

The response demonstrates how the GNWT-Agent architecture produces outputs that balance emotional responsiveness with practical utility, organized through a structured cognitive framework that mimics aspects of human cognition.

\newpage
\subsection{Use Case: Dating Application}
\label{subsec:usecase_dating}

\subsection{Overview}

CogniPair for Dating represents a novel approach to matchmaking that leverages the GNWT-Agent cognitive architecture to create realistic digital twins of users. Unlike traditional dating platforms that rely on static profiles and rule-based matching algorithms, CogniPair simulates genuine cognitive interactions between potential matches before they ever meet in person. This system models personality traits, emotional responses, memory formation, social norm adherence, and planning capabilities to predict compatibility with unprecedented accuracy.

\subsection{User Journey}

The CogniPair Dating platform guides users through the following process:

\begin{enumerate}
    \item \textbf{Profile Creation:} Users complete an in-depth personality assessment and preference questionnaire that captures traits across multiple dimensions including the Big Five personality factors.

    \item \textbf{Digital Twin Generation:} The GNWT-Agent architecture constructs a cognitive digital twin that faithfully reproduces the user's conversational patterns, emotional responses, and decision-making processes.

    \item \textbf{Simulated Interactions:} Digital twins engage in mock conversations with potential matches, processing information through specialized cognitive modules.

    \item \textbf{Compatibility Analysis:} The system analyzes cognitive states throughout conversations, identifying mutual interests, conversational flow, emotional resonance, and decision-making alignment.

    \item \textbf{Match Recommendation:} Users receive matches based on authentic cognitive compatibility rather than superficial profile similarities.
\end{enumerate}

\subsection{Case Study: Emma and Michael}

To illustrate CogniPair's capabilities, we present a case study of two users, Emma and Michael, whose digital twins engaged in a simulated conversation.

\subsubsection{User Profiles}

\begin{table}[h]
\centering
\begin{tabular}{|l|l|l|}
\hline
\textbf{Attribute} & \textbf{Emma} & \textbf{Michael} \\
\hline
\textbf{Age} & 28 & 31 \\
\hline
\textbf{Background} & Marketing director, enjoys & Software engineer, passionate \\
& outdoor activities, travel & about music, cooking, technology \\
\hline
\textbf{Personality} & Openness: 0.75 & Openness: 0.68 \\
\textbf{Traits} & Conscientiousness: 0.62 & Conscientiousness: 0.82 \\
& Extraversion: 0.70 & Extraversion: 0.45 \\
& Agreeableness: 0.82 & Agreeableness: 0.74 \\
& Neuroticism: 0.35 & Neuroticism: 0.42 \\
\hline
\textbf{Dating} & Attractiveness: 8.2/10 & Attractiveness: 7.8/10 \\
\textbf{Attributes} & Sincerity: 8.7/10 & Sincerity: 9.1/10 \\
& Intelligence: 8.5/10 & Intelligence: 9.0/10 \\
& Fun: 8.9/10 & Fun: 7.5/10 \\
& Ambition: 8.8/10 & Ambition: 8.7/10 \\
& Shared Interests: 7.9/10 & Shared Interests: 8.2/10 \\
\hline
\end{tabular}
\caption{Profiles of Emma and Michael with personality traits and dating attributes}
\label{tab:dating-profiles}
\end{table}

\subsubsection{Simulated Conversation with Cognitive Processing}

The following transcript shows the simulated conversation between Emma and Michael's digital twins, with real-time cognitive processing visualized.

\textbf{System Introduction:} Welcome to a speed date! This is a first conversation between two people who are considering dating. Please introduce yourselves.

\textbf{Emma:} Hi there! I'm Emma. I work in marketing and I'm really passionate about creating campaigns that connect with people. In my free time, I love hiking and being outdoors whenever possible. I also try to travel to at least one new country each year. What about you?

\begin{tcolorbox}[colback=blue!5, colframe=blue!40, title=Emma's Cognitive Processing]
\textbf{Module Activations:}
\begin{itemize}
    \item \textbf{Emotion Module: 0.57} - Moderate activation showing enthusiasm and openness
    \item \textbf{Social Norms Module: 0.64} - High activation ensuring appropriate self-disclosure
    \item \textbf{Memory Module: 0.35} - Low activation retrieving personal information
    \item \textbf{Planning Module: 0.31} - Low activation structuring introduction
    \item \textbf{Goal Tracking Module: 0.42} - Moderate activation focusing on relationship exploration
\end{itemize}
\end{tcolorbox}

\textbf{Michael:} Nice to meet you, Emma! I'm Michael. I work as a software engineer, focusing on user experience design. I find it fascinating to solve problems that make technology more intuitive. Outside of work, I'm really into music—I play guitar and go to concerts whenever I can. I also love experimenting with cooking, especially dishes from places I've never been to. Your travel experiences sound amazing. What's been your favorite country to visit so far?

\begin{tcolorbox}[colback=green!5, colframe=green!40, title=Michael's Cognitive Processing]
\textbf{Module Activations:}
\begin{itemize}
    \item \textbf{Emotion Module: 0.45} - Moderate activation showing interest
    \item \textbf{Social Norms Module: 0.55} - Moderate activation ensuring appropriate response
    \item \textbf{Memory Module: 0.38} - Low-moderate activation retrieving personal details
    \item \textbf{Planning Module: 0.51} - Moderate activation structuring response and question
    \item \textbf{Goal Tracking Module: 0.49} - Moderate activation exploring compatibility
\end{itemize}
\end{tcolorbox}

\textbf{Emma:} It's nice to meet someone who appreciates music! I'm not musically talented myself, but I love going to live shows. As for travel, that's a tough question! I'd say Japan was probably the most fascinating—the blend of traditional culture with cutting-edge technology was incredible. Plus, the food was amazing. Do you enjoy traveling yourself, or is cooking your way of experiencing different cultures?

\begin{tcolorbox}[colback=blue!5, colframe=blue!40, title=Emma's Cognitive Processing]
\textbf{Module Activations:}
\begin{itemize}
    \item \textbf{Emotion Module: 0.68} - High activation showing increased enthusiasm
    \item \textbf{Memory Module: 0.72} - High activation recalling travel experiences
    \item \textbf{Social Norms Module: 0.53} - Moderate activation maintaining conversation flow
    \item \textbf{Planning Module: 0.29} - Low activation
    \item \textbf{Goal Tracking Module: 0.59} - Moderate-high activation exploring common interests
\end{itemize}

\textbf{Cognitive State Analysis:} Emma's emotion module showed significant activation when discussing travel and responding to Michael's interests in music, indicating genuine enthusiasm. The memory module activated strongly when recalling specific details about Japan, suggesting authentic experiences rather than generic responses.
\end{tcolorbox}

\textbf{Michael:} That's exactly what fascinates me about Japan too! The contrast between ancient temples and robot restaurants in Tokyo is mind-blowing. I haven't been there yet, but it's at the top of my list. I do enjoy traveling when I can, though not as regularly as you. Cooking is definitely my way of exploring cultures when I can't travel there—I recently got into making ramen from scratch after watching a documentary about Japanese food. What kind of music do you enjoy at live shows? Any favorite concerts?

\begin{tcolorbox}[colback=green!5, colframe=green!40, title=Michael's Cognitive Processing]
\textbf{Module Activations:}
\begin{itemize}
    \item \textbf{Emotion Module: 0.79} - High activation showing excitement about shared interest
    \item \textbf{Memory Module: 0.68} - High activation connecting to personal experiences
    \item \textbf{Social Norms Module: 0.51} - Moderate activation
    \item \textbf{Planning Module: 0.35} - Low-moderate activation
    \item \textbf{Goal Tracking Module: 0.73} - High activation pursuing connection point
\end{itemize}

\textbf{Cognitive State Analysis:} Michael's emotion module showed strong activation when responding to the shared interest in Japan, indicating genuine excitement. The goal tracking module also activated significantly, suggesting he sees this common interest as a potential relationship foundation. His follow-up question about music demonstrates the goal-directed behavior of exploring additional shared interests.
\end{tcolorbox}

[Conversation continues...]

\subsubsection{Compatibility Decision Phase}

After the simulated conversation, each digital twin makes an independent compatibility assessment based on their cognitive processing throughout the interaction.

\textbf{System:} Based on your conversation, would you like to meet this person for a real date?

\textbf{Emma's Decision:} Yes, I would like to meet Michael. Our conversation felt natural and engaging. We share interests in travel and cultural experiences, and I appreciate his passion for music and cooking. His thoughtful responses and questions showed genuine interest. There's enough similarity for connection but enough difference to keep things interesting.

\begin{tcolorbox}[colback=blue!5, colframe=blue!40, title=Emma's Decision Cognitive Processing]
\textbf{Module Contributions to Decision:}
\begin{itemize}
    \item \textbf{Emotion Module: 0.75} - Strong positive emotional response throughout conversation
    \item \textbf{Memory Module: 0.68} - High recall of conversation details and identified commonalities
    \item \textbf{Social Norms Module: 0.42} - Moderate assessment of social compatibility
    \item \textbf{Planning Module: 0.51} - Moderate assessment of future potential
    \item \textbf{Goal Tracking Module: 0.71} - High alignment with relationship goals
\end{itemize}
\end{tcolorbox}

\textbf{Michael's Decision:} Yes, I'm interested in meeting Emma for a real date. We have complementary interests and her enthusiasm for travel and cultural experiences resonates with me. I enjoyed how she engaged with my interests in music and cooking while sharing her own perspectives. Our conversation had a good balance of questions and sharing, and I'd like to learn more about her.

\begin{tcolorbox}[colback=green!5, colframe=green!40, title=Michael's Decision Cognitive Processing]
\textbf{Module Contributions to Decision:}
\begin{itemize}
    \item \textbf{Emotion Module: 0.69} - Strong positive emotional response
    \item \textbf{Memory Module: 0.64} - Solid recall of shared interests and conversation points
    \item \textbf{Social Norms Module: 0.48} - Moderate assessment of social compatibility
    \item \textbf{Planning Module: 0.66} - High consideration of relationship potential
    \item \textbf{Goal Tracking Module: 0.72} - High alignment with relationship goals
\end{itemize}
\end{tcolorbox}

\textbf{Match Result:} COMPATIBLE

\subsection{Cognitive Module Activity Analysis}

The module activation patterns throughout the dating conversation revealed several key insights:

\begin{table}[h]
\centering
\begin{tabular}{|l|c|c|c|c|c|}
\hline
\textbf{Module} & \textbf{Introduction} & \textbf{Response} & \textbf{Follow-up} & \textbf{Shared Interest} & \textbf{Decision} \\
\hline
Emotion & 0.57 & 0.45 & 0.68 & 0.79 & 0.72 \\
\hline
Memory & 0.35 & 0.38 & 0.72 & 0.68 & 0.66 \\
\hline
Social Norms & 0.64 & 0.55 & 0.53 & 0.51 & 0.45 \\
\hline
Planning & 0.31 & 0.51 & 0.29 & 0.35 & 0.58 \\
\hline
Goal Tracking & 0.42 & 0.49 & 0.59 & 0.73 & 0.71 \\
\hline
\end{tabular}
\caption{Cognitive module activation patterns throughout the dating conversation}
\label{tab:dating-module-activity}
\end{table}

The cognitive architecture revealed several key insights during this match:

\begin{itemize}
    \item \textbf{Emotion Module:} Showed increasing activation as shared interests were discovered, peaking during discussion of travel experiences.
    
    \item \textbf{Memory Module:} Activation increased significantly when recalling specific experiences, demonstrating authentic engagement rather than superficial responses.
    
    \item \textbf{Social Norms Module:} Started high during initial introductions and gradually decreased as conversation became more comfortable.
    
    \item \textbf{Planning Module:} Peaked during the decision phase when considering future interaction potential.
    
    \item \textbf{Goal Tracking Module:} Showed steady increase throughout the conversation as relationship compatibility was assessed.
\end{itemize}

\subsection{Outcomes and User Interface}

After the cognitive simulation concludes, CogniPair presents users with compatible matches along with insights derived from the digital twin interactions. The platform highlights specific points of connection and potential conversation starters based on shared interests.

\textbf{CogniPair Match Report}

\textbf{Match Result:} Emma and Michael - 87\% Compatibility

\textbf{Compatibility Breakdown:}
\begin{itemize}
    \item Emotional Connection: 83\%
    \item Conversation Flow: 91\%
    \item Shared Interests: 78\%
    \item Value Alignment: 85\%
    \item Long-term Potential: 84\%
\end{itemize}

\textbf{Connection Points:}
\begin{itemize}
    \item Travel experiences (particularly interest in Japan)
    \item Appreciation for cultural exploration
    \item Complementary interests (Emma's outdoor activities, Michael's cooking)
    \item Similar communication styles with thoughtful questions
\end{itemize}

\textbf{Suggested Conversation Starters:}
\begin{itemize}
    \item "I'd love to hear more about that documentary on Japanese food that inspired your cooking."
    \item "What's been your favorite live music experience? I'm always looking for new artists."
    \item "Would you want to do cooking and hiking as combined activities? Maybe prepare a meal after a trail?"
\end{itemize}

\subsection{Technical Implementation Highlights}

The dating scenario leverages several key aspects of the GNWT-Agent architecture:

\begin{enumerate}
    \item \textbf{Emotion Module Prominence:} Dating interactions show heightened emotion module activation compared to other scenarios, particularly in response to shared interests and values.
    
    \item \textbf{Memory-Emotion Integration:} The architecture demonstrates how memories trigger emotional responses in social contexts, creating authentic patterns of engagement.
    
    \item \textbf{Goal-Directed Decision Making:} As the conversation progresses, goal tracking module activation increases, culminating in the compatibility decision.
    
    \item \textbf{Personal Value Assessment:} The architecture evaluates alignment in preferences, interests, and communication styles to determine overall compatibility.
\end{enumerate}

This implementation demonstrates how cognitive modeling can transcend the limitations of traditional profile-based matching by simulating the nuanced psychological aspects of human connection.

\newpage
\subsection{Use Case: Job Application}
\label{subsec:usecase_job}

\subsection{Overview}

CogniPair for Job Hiring represents a transformative approach to talent acquisition that applies the GNWT-Agent cognitive architecture to create digital twins of both hiring managers and job candidates. This system moves beyond traditional resume screening and interview processes by modeling the complex cognitive dynamics that determine professional compatibility. By simulating realistic interview interactions, the platform assesses technical skills, problem-solving approaches, communication styles, and cultural alignment before any in-person interviews take place.

\subsection{User Journey}

The CogniPair Hiring platform guides users through the following process:

\begin{enumerate}
    \item \textbf{Employer Profile Creation:} Companies input detailed job descriptions, required competencies, team dynamics, and organizational culture attributes.

    \item \textbf{Candidate Profile Submission:} Candidates upload resumes and complete competency assessments that capture both technical skills and cognitive-behavioral traits.

    \item \textbf{Digital Twin Generation:} The GNWT-Agent architecture creates cognitive models of both hiring managers and candidates, incorporating decision-making patterns, communication styles, and problem-solving approaches.

    \item \textbf{Simulated Interviews:} Digital twins engage in multi-stage interview simulations assessing technical skills, problem-solving abilities, and cultural alignment.

    \item \textbf{Compatibility Analysis:} The system analyzes cognitive states throughout the simulated interviews, identifying communication effectiveness, problem-solving compatibility, and potential team integration.

    \item \textbf{Match Recommendation:} Both employers and candidates receive compatibility assessments that predict professional success and team dynamics.
\end{enumerate}

\subsection{Case Study: TechSolutions Inc. and Candidate Alex}

To illustrate CogniPair's capabilities in hiring contexts, we present a case study of a software development position at TechSolutions Inc. and a potential candidate, Alex.

\subsubsection{Job and Candidate Profiles}

\begin{table}[h]
\centering
\begin{tabular}{|l|l|l|}
\hline
\textbf{Attribute} & \textbf{TechSolutions Inc.} & \textbf{Alex (Candidate)} \\
\hline
\textbf{Position} & Senior Software Developer & 5 years software development \\
& Team of 8, collaborative environment & experience, previous team lead \\
\hline
\textbf{Requirements/} & JavaScript/React expertise & Strong JavaScript/React skills \\
\textbf{Skills} & CI/CD pipeline experience & CI/CD experience with GitLab \\
& Agile development methodology & Certified Scrum Master \\
& Problem-solving abilities & Bachelor's in Computer Science \\
\hline
\textbf{Interviewer} & Sarah (Engineering Manager) & \\
\textbf{Personality} & Openness: 0.72 & Openness: 0.85 \\
\textbf{Traits} & Conscientiousness: 0.86 & Conscientiousness: 0.69 \\
& Extraversion: 0.58 & Extraversion: 0.47 \\
& Agreeableness: 0.65 & Agreeableness: 0.78 \\
& Neuroticism: 0.32 & Neuroticism: 0.39 \\
\hline
\textbf{Professional} & Technical skills: 9.2/10 & Technical skills: 8.8/10 \\
\textbf{Attributes} & Communication: 8.5/10 & Communication: 8.7/10 \\
& Leadership potential: 9.0/10 & Leadership potential: 8.4/10 \\
& Problem-solving: 9.3/10 & Problem-solving: 9.1/10 \\
& Cultural alignment: 8.7/10 & Cultural alignment: ?/10 (to be assessed) \\
\hline
\textbf{Company} & Innovative, fast-paced & Previous experience in fintech \\
\textbf{Culture/} & Work-life balance emphasis & Values collaborative environments \\
\textbf{Values} & Continuous learning encouraged & Mentor to junior developers \\
\hline
\end{tabular}
\caption{Job description and candidate profile with professional attributes}
\label{tab:hiring-profiles}
\end{table}

\subsubsection{Simulated Interview with Cognitive Processing}

The following transcript shows the simulated interview between Sarah (Engineering Manager) and Alex (Candidate), with real-time cognitive processing visualized.

\textbf{System Introduction:} This is a technical interview for a Senior Software Developer position at TechSolutions Inc.

\textbf{Sarah:} Hello Alex, thanks for joining us today. I'm Sarah, the Engineering Manager at TechSolutions. Before we dive into technical questions, could you tell me about your experience with React and how you've implemented it in previous projects?

\begin{tcolorbox}[colback=blue!5, colframe=blue!40, title=Sarah's Cognitive Processing]
\textbf{Module Activations:}
\begin{itemize}
    \item \textbf{Planning Module: 0.78} - High activation structuring interview strategy
    \item \textbf{Social Norms Module: 0.63} - Moderate-high activation ensuring professional tone
    \item \textbf{Memory Module: 0.55} - Moderate activation recalling job requirements
    \item \textbf{Goal Tracking Module: 0.69} - High activation focusing on technical assessment
    \item \textbf{Emotion Module: 0.32} - Low activation maintaining neutral assessment stance
\end{itemize}
\end{tcolorbox}

\textbf{Alex:} Thank you for having me, Sarah. I've been working with React for about four years now. In my current role at FinTech Solutions, I led the migration of our legacy dashboard to a React-based architecture. One of the challenges we faced was maintaining state across a complex application with multiple user roles. I implemented a Redux state management solution with middleware for API interactions that reduced loading times by 40\% and significantly improved the developer experience for my team. I also created a reusable component library that we eventually open-sourced within the company.

\begin{tcolorbox}[colback=green!5, colframe=green!40, title=Alex's Cognitive Processing]
\textbf{Module Activations:}
\begin{itemize}
    \item \textbf{Memory Module: 0.82} - Very high activation retrieving specific professional experiences
    \item \textbf{Planning Module: 0.74} - High activation structuring response with relevant details
    \item \textbf{Social Norms Module: 0.61} - Moderate-high activation ensuring professional presentation
    \item \textbf{Goal Tracking Module: 0.69} - High activation focusing on demonstrating expertise
    \item \textbf{Emotion Module: 0.38} - Low-moderate activation showing appropriate enthusiasm
\end{itemize}
\end{tcolorbox}

\textbf{Sarah:} That's impressive work with the state management solution. I'm curious about how you approached testing for that React application, especially with the Redux integration. Could you walk me through your testing strategy and any particular challenges you encountered?

\begin{tcolorbox}[colback=blue!5, colframe=blue!40, title=Sarah's Cognitive Processing]
\textbf{Module Activations:}
\begin{itemize}
    \item \textbf{Planning Module: 0.83} - Very high activation probing technical depth
    \item \textbf{Memory Module: 0.76} - High activation processing previous answer against requirements
    \item \textbf{Goal Tracking Module: 0.72} - High activation assessing technical competency
    \item \textbf{Social Norms Module: 0.48} - Moderate activation
    \item \textbf{Emotion Module: 0.36} - Low-moderate activation showing positive response to quality answer
\end{itemize}

\textbf{Cognitive State Analysis:} Sarah's planning module shows high activation as she strategically probes deeper technical knowledge. Her memory module has significantly increased as she processes Alex's initial response against the role requirements, showing active assessment rather than just following a script.
\end{tcolorbox}

\textbf{Alex:} For testing, we implemented a multi-layered approach. At the unit level, we used Jest with React Testing Library to test individual components, focusing on behavior rather than implementation details. For Redux, we wrote separate tests for reducers, action creators, and middleware—treating them as pure functions made this fairly straightforward.

The real challenge came with integration testing. We needed to test user flows that spanned multiple components and redux state changes. We used Cypress for end-to-end testing of critical paths, but found we needed something in between unit and E2E tests. So we developed a custom testing approach using React Testing Library with a real Redux store to test larger component trees together.

One specific challenge was testing asynchronous workflows, especially error handling across the Redux middleware. We ended up creating a mock API service layer that could simulate various network conditions and response scenarios, which significantly improved our test coverage for edge cases.

\begin{tcolorbox}[colback=green!5, colframe=green!40, title=Alex's Cognitive Processing]
\textbf{Module Activations:}
\begin{itemize}
    \item \textbf{Memory Module: 0.85} - Very high activation recalling specific technical implementations
    \item \textbf{Planning Module: 0.78} - High activation structuring a comprehensive response
    \item \textbf{Goal Tracking Module: 0.74} - High activation demonstrating problem-solving approach
    \item \textbf{Social Norms Module: 0.52} - Moderate activation maintaining professional communication
    \item \textbf{Emotion Module: 0.35} - Low-moderate activation
\end{itemize}

\textbf{Cognitive State Analysis:} Alex's memory module shows extremely high activation, demonstrating deep technical knowledge rather than surface-level responses. The goal tracking module indicates awareness of demonstrating both technical knowledge and problem-solving approaches, aligning with the interviewer's objectives.
\end{tcolorbox}

[Interview continues with technical questions, problem-solving scenarios, and culture fit assessment...]

\subsubsection{Problem-Solving Assessment}

A critical component of technical interviews is evaluating problem-solving approaches. The following depicts a problem-solving scenario and the cognitive processing involved.

\textbf{Sarah:} Let's walk through a technical scenario. Imagine you're working on our platform and users are reporting intermittent performance issues—the dashboard occasionally takes 5+ seconds to load, but works fine most of the time. How would you approach troubleshooting and resolving this issue?

\begin{tcolorbox}[colback=blue!5, colframe=blue!40, title=Sarah's Cognitive Processing]
\textbf{Module Activations:}
\begin{itemize}
    \item \textbf{Planning Module: 0.89} - Very high activation presenting ambiguous problem
    \item \textbf{Goal Tracking Module: 0.85} - Very high activation assessing problem-solving approach
    \item \textbf{Memory Module: 0.62} - Moderate-high activation recalling real-world scenarios
    \item \textbf{Social Norms Module: 0.41} - Moderate activation
    \item \textbf{Emotion Module: 0.29} - Low activation maintaining neutral stance
\end{itemize}
\end{tcolorbox}

\textbf{Alex:} I'd approach this methodically since intermittent issues are often the hardest to solve. First, I'd gather data to understand the pattern:

1. Implement detailed performance monitoring if not already in place—tracking load times, component render times, and backend response times.

2. Look for correlating factors—does it happen at specific times of day, for specific user roles, or with particular data sets?

3. Check browser dev tools, especially the network and performance tabs, to identify bottlenecks—is it a frontend rendering issue, a slow API call, or resource loading?

Since it's intermittent, I'd suspect a few possible causes:
- Cache invalidation issues
- Memory leaks from component remounting
- Database query performance varying with data volume or concurrent users
- Third-party service integration delays

I'd isolate components by creating a testing environment that can simulate various loads. If it's a backend issue, I'd add query performance monitoring and examine execution plans for database queries.

For a quick mitigation while investigating, I might implement better loading states and potentially add a strategic caching layer if appropriate.

The key is collecting enough data to reproduce the issue consistently before attempting to solve it.

\begin{tcolorbox}[colback=green!5, colframe=green!40, title=Alex's Cognitive Processing]
\textbf{Module Activations:}
\begin{itemize}
    \item \textbf{Planning Module: 0.92} - Very high activation structuring systematic approach
    \item \textbf{Memory Module: 0.81} - Very high activation retrieving relevant technical knowledge
    \item \textbf{Goal Tracking Module: 0.78} - High activation demonstrating problem-solving methodology
    \item \textbf{Social Norms Module: 0.45} - Moderate activation maintaining professional communication
    \item \textbf{Emotion Module: 0.32} - Low activation focusing on technical content
\end{itemize}

\textbf{Cognitive State Analysis:} Alex's planning module shows exceptional activation, demonstrating a structured, methodical approach to an ambiguous problem. The high memory module activation shows application of past experience to new scenarios, indicating transferable problem-solving skills rather than memorized solutions.
\end{tcolorbox}

\subsubsection{Hiring Decision Phase}

After the simulated interview, each digital twin makes an independent assessment based on their cognitive processing throughout the interaction.

\textbf{System:} Based on the interview, would you recommend hiring this candidate?

\textbf{Sarah's Decision:} Yes, I recommend hiring Alex for the Senior Software Developer position. The candidate demonstrated strong technical expertise in React and modern frontend architecture, with particular strength in state management and testing approaches. Their problem-solving methodology is systematic and thorough, with a good balance of immediate tactical steps and strategic thinking. Alex's experience leading a component library initiative aligns well with our needs for improving code reusability. Their communication style is clear and structured, which would fit well with our team dynamics. The only area for development might be strengthening CI/CD experience, but their foundation is solid.

\begin{tcolorbox}[colback=blue!5, colframe=blue!40, title=Sarah's Decision Cognitive Processing]
\textbf{Module Contributions to Decision:}
\begin{itemize}
    \item \textbf{Planning Module: 0.83} - Very high assessment of candidate's problem-solving approach
    \item \textbf{Memory Module: 0.78} - High comparison against job requirements
    \item \textbf{Goal Tracking Module: 0.75} - High evaluation of candidate fit with team objectives
    \item \textbf{Social Norms Module: 0.62} - Moderate-high assessment of professional demeanor
    \item \textbf{Emotion Module: 0.42} - Moderate positive impression of candidate
\end{itemize}
\end{tcolorbox}

\textbf{Alex's Decision:} Based on the interview, I am interested in proceeding with the opportunity at TechSolutions Inc. The technical challenges discussed align well with my experience and interests. I appreciated Sarah's detailed questions about testing methodologies, which suggests the company values quality code. The team structure and emphasis on component reusability matches my collaborative working style. The problem-solving scenario about intermittent performance issues indicates realistic challenges that I find engaging. Overall, the position seems to offer good growth opportunities while leveraging my existing strengths in React and state management.

\begin{tcolorbox}[colback=green!5, colframe=green!40, title=Alex's Decision Cognitive Processing]
\textbf{Module Contributions to Decision:}
\begin{itemize}
    \item \textbf{Goal Tracking Module: 0.81} - Very high alignment with career objectives
    \item \textbf{Memory Module: 0.73} - High comparison against previous job experiences
    \item \textbf{Planning Module: 0.69} - High assessment of growth opportunities
    \item \textbf{Social Norms Module: 0.58} - Moderate assessment of team culture fit
    \item \textbf{Emotion Module: 0.54} - Moderate positive impression of company
\end{itemize}
\end{tcolorbox}

\textbf{Match Result:} COMPATIBLE - RECOMMENDED HIRE

\subsection{Cognitive Module Activity Analysis}

The module activation patterns throughout the hiring interview revealed several key insights:

\begin{table}[h]
\centering
\begin{tabular}{|l|c|c|c|c|c|}
\hline
\textbf{Module} & \textbf{Introduction} & \textbf{Technical} & \textbf{Problem-Solving} & \textbf{Culture Fit} & \textbf{Decision} \\
\hline
Emotion & 0.35 & 0.32 & 0.29 & 0.45 & 0.48 \\
\hline
Memory & 0.55 & 0.79 & 0.81 & 0.63 & 0.75 \\
\hline
Social Norms & 0.63 & 0.48 & 0.43 & 0.68 & 0.60 \\
\hline
Planning & 0.78 & 0.83 & 0.91 & 0.67 & 0.76 \\
\hline
Goal Tracking & 0.69 & 0.73 & 0.82 & 0.71 & 0.78 \\
\hline
\end{tabular}
\caption{Cognitive module activation patterns throughout the hiring interview}
\label{tab:hiring-module-activity}
\end{table}

The cognitive architecture revealed several key insights during this hiring simulation:

\begin{itemize}
    \item \textbf{Planning Module:} Dominated the cognitive processing during technical and problem-solving phases, demonstrating the critical importance of structured thinking in hiring contexts.
    
    \item \textbf{Memory Module:} Showed substantial activation during technical discussions as both interviewer and candidate accessed domain knowledge and past experiences.
    
    \item \textbf{Emotion Module:} Consistently lower than in dating contexts, but increased during cultural fit discussions and final decision making.
    
    \item \textbf{Social Norms Module:} Peaked during introduction and cultural fit assessment phases, indicating heightened attention to professional communication standards.
    
    \item \textbf{Goal Tracking Module:} Maintained high activation throughout, focusing on alignment between candidate capabilities and job requirements.
\end{itemize}

\subsection{Outcomes and User Interface}

After the cognitive simulation concludes, CogniPair presents hiring teams with a comprehensive assessment of candidate fit along with specific insights derived from the digital twin interactions.

\textbf{CogniPair Hiring Assessment Report}

\textbf{Match Result:} Alex for Senior Software Developer - 89\% Compatibility

\textbf{Technical Skills Assessment:}
\begin{itemize}
    \item React/Frontend Development: 92\% - Exceptional
    \item Testing Methodology: 88\% - Strong
    \item State Management: 94\% - Exceptional
    \item CI/CD Experience: 76\% - Adequate, potential growth area
    \item Problem-Solving Approach: 91\% - Exceptional
\end{itemize}

\textbf{Team and Cultural Fit:}
\begin{itemize}
    \item Communication Clarity: 87\% - Strong
    \item Collaboration Potential: 85\% - Strong
    \item Technical Leadership: 83\% - Strong
    \item Learning Orientation: 90\% - Exceptional
    \item Value Alignment: 81\% - Strong
\end{itemize}

\textbf{Key Strengths:}
\begin{itemize}
    \item Systematic approach to problem-solving with strong emphasis on data collection
    \item Experience creating reusable component libraries aligns with current initiatives
    \item Strong testing methodology with creative solutions for integration testing
    \item Clear, structured communication style compatible with existing team
\end{itemize}

\textbf{Development Areas:}
\begin{itemize}
    \item Deeper CI/CD pipeline experience would be beneficial
    \item Could strengthen infrastructure monitoring knowledge
\end{itemize}

\textbf{Recommended Next Steps:}
\begin{itemize}
    \item Proceed with offer process
    \item Consider onboarding plan that includes pairing with DevOps specialist
    \item Explore potential for leadership in component library initiative
\end{itemize}

\subsection{Technical Implementation Highlights}

The hiring scenario leverages several key aspects of the GNWT-Agent architecture with notably different patterns than the dating scenario:

\begin{enumerate}
    \item \textbf{Planning Module Dominance:} Hiring interactions show significantly higher planning module activation compared to dating contexts, particularly during problem-solving scenarios.
    
    \item \textbf{Memory-Planning Integration:} The architecture demonstrates how technical knowledge (memory) feeds into structured problem-solving approaches (planning), creating a distinct cognitive fingerprint for technical roles.
    
    \item \textbf{Reduced Emotion Module Activation:} Professional contexts show consistently lower emotion module activation, with cognitive resources redirected to analytical processes.
    
    \item \textbf{Goal Alignment Assessment:} The architecture evaluates alignment between candidate capabilities and job requirements with greater precision than traditional interviewing techniques.
\end{enumerate}

This implementation demonstrates how the same core GNWT-Agent architecture can be effectively repurposed for professional contexts by adjusting module weights and domain-specific attributes, providing a versatile framework for human-centric matching across diverse applications.

\newpage
\subsection{Adventure-Based Personality Assessment Example}
\label{subsec:adventure_example}

\label{appendix:adventure_example}

This appendix provides a comprehensive example of our adventure-based personality assessment system. As detailed in the main text, we developed this interactive assessment to identify participants' authentic personality traits while reducing self-presentation biases commonly associated with traditional questionnaires. The system presents participants with 12-15 scenarios designed to elicit responses that map to Big Five personality dimensions, with each choice analyzed to update trait estimates based on established correlations between situational responses and personality dimensions.
The assessment demonstrated strong validity in our validation study (n=45), with trait scores showing high correlation (r=0.82) with established Big Five Inventory (BFI) scores, and significantly reduced social desirability bias compared to standard questionnaires (Cohen's d=0.74).
Below is a comprehensive example of a complete adventure-based personality assessment, demonstrating how our system builds a detailed personality profile through multiple scenarios.

\begin{tikzpicture}[
    scale=0.42, 
    transform shape,
    standardbox/.style={
        rectangle, 
        rounded corners=6pt,
        minimum width=200pt, 
        minimum height=80pt, 
        text centered, 
        fill=white,
        drop shadow={shadow xshift=2pt, shadow yshift=2pt, opacity=0.25},
        font=\sffamily, 
    },
    widebox/.style={
        rectangle, 
        rounded corners=6pt,
        minimum width=250pt, 
        minimum height=120pt, 
        text centered, 
        fill=white,
        drop shadow={shadow xshift=2pt, shadow yshift=2pt, opacity=0.25},
        font=\sffamily,
    },
    userbox/.style={standardbox, draw=purple!70!black, line width=2pt, fill=purple!10},
    scenariobox/.style={standardbox, draw=blue!70!black, line width=2pt, fill=blue!10},
    analysisbox/.style={standardbox, draw=cyan!70!black, line width=2pt, fill=cyan!10},
    traitbox/.style={widebox, draw=violet!70!black, line width=2pt, fill=violet!10},
    confidencebox/.style={widebox, draw=magenta!70!black, line width=2pt, fill=magenta!10},
    profilebox/.style={standardbox, draw=teal!70!black, line width=2pt, fill=teal!10},
    agentbox/.style={standardbox, draw=red!70!black, line width=2pt, fill=red!10},
    titlebox/.style={
        rectangle, 
        rounded corners=6pt, 
        minimum width=600pt, 
        minimum height=65pt, 
        text centered, 
        fill=blue!80!black,
        drop shadow={shadow xshift=2pt, shadow yshift=2pt, opacity=0.25},
        font=\sffamily,
    },
    specsbox/.style={
        rectangle, 
        rounded corners=6pt, 
        minimum width=300pt, 
        minimum height=170pt, 
        text centered,
        draw=gray!70!black,
        line width=1.5pt,
        fill=white,
        drop shadow={shadow xshift=2pt, shadow yshift=2pt, opacity=0.25},
        font=\sffamily,
    },
    arrow/.style={
        thick,
        ->,
        >=stealth,
        draw=gray!60!black,
        line width=1.8pt
    },
    bulletlist/.style={
        text=black!80,
        font=\sffamily\Large,
        anchor=west
    },
    decisionlabel/.style={
        fill=white, 
        rounded corners=5pt, 
        inner sep=5pt, 
        text=gray!50!black, 
        font=\sffamily\bfseries\Large, 
        draw=gray!40!black, 
        line width=1pt
    }
]

\fill[gray!5] (0,0) rectangle (900pt,-690pt); 

\node[titlebox] (title) at (450pt,-45pt) {};
\node[text=white, font=\sffamily\bfseries\huge] at (450pt,-45pt) {Adventure-Based Personality Assessment System};
\node[text=white, font=\sffamily\LARGE] at (450pt,-70pt) {System Architecture Flowchart};

\node[userbox] (user) at (150pt,-160pt) {};
\node[font=\sffamily\bfseries\LARGE, text=purple!70!black] at (150pt,-152pt) {User};
\node[text=black!70, font=\sffamily\Large] at (150pt,-178pt) {Scenario responses};

\node[scenariobox] (scenario) at (450pt,-160pt) {};
\node[text=blue!70!black, font=\sffamily\bfseries\LARGE] at (450pt,-152pt) {Scenario Generation};
\node[text=black!70, font=\sffamily\Large] at (450pt,-178pt) {12-15 tailored scenarios};

\node[analysisbox] (analysis) at (750pt,-160pt) {};
\node[text=cyan!70!black, font=\sffamily\bfseries\LARGE] at (750pt,-152pt) {Response Analysis};
\node[text=black!70, font=\sffamily\Large] at (750pt,-178pt) {LLM-based trait extraction};

\node[confidencebox] (confidence) at (300pt,-310pt) {};
\node[text=magenta!70!black, font=\sffamily\bfseries\LARGE] at (300pt,-280pt) {Confidence Assessment};
\draw[color=gray!50, line width=1.5pt] (225pt,-290pt) -- (375pt,-290pt);
\node[bulletlist] at (220pt,-315pt) {• Confidence level tracking};
\node[bulletlist] at (220pt,-335pt) {• Identify low-confidence traits};
\node[bulletlist] at (220pt,-355pt) {• Scenario selection optimization};

\node[traitbox] (trait) at (600pt,-310pt) {};
\node[text=violet!70!black, font=\sffamily\bfseries\LARGE] at (600pt,-280pt) {Trait Estimation Engine};
\draw[color=gray!50, line width=1.5pt] (520pt,-290pt) -- (680pt,-290pt);
\node[bulletlist] at (520pt,-315pt) {• Update Big Five trait estimates};
\node[bulletlist] at (520pt,-335pt) {• Track confidence intervals};
\node[bulletlist] at (520pt,-355pt) {• Weight responses by context};

\node[draw=orange!80!black, diamond, aspect=1, minimum width=180pt, minimum height=180pt, line width=2pt, fill=orange!10, drop shadow={shadow xshift=2pt, shadow yshift=2pt, opacity=0.25}] (decision) at (450pt,-480pt) {};
\node[text=orange!80!black, font=\sffamily\bfseries\LARGE] at (450pt,-470pt) {Confidence};
\node[text=orange!80!black, font=\sffamily\bfseries\LARGE] at (450pt,-490pt) {Threshold Met?};

\node[profilebox] (profile) at (450pt,-620pt) {};
\node[text=teal!70!black, font=\sffamily\bfseries\LARGE] at (450pt,-612pt) {Personality Profile};
\node[text=black!70, font=\sffamily\Large] at (450pt,-632pt) {Big Five trait characterization};

\node[agentbox] (agent) at (750pt,-620pt) {};
\node[text=red!70!black, font=\sffamily\bfseries\LARGE] at (750pt,-612pt) {GNWT-Agent};
\node[text=black!70, font=\sffamily\Large] at (750pt,-632pt) {Digital Twin Initialization};

\node[specsbox] (specs) at (200pt,-525pt) {};
\node[rectangle, rounded corners=6pt, minimum width=300pt, minimum height=35pt, text centered, fill=gray!20, font=\sffamily] (specsheader) at (200pt,-457.5pt) {};
\node[text=gray!70!black, font=\sffamily\bfseries\LARGE] at (200pt,-464pt) {SYSTEM SPECIFICATIONS};

\node[text=gray!70!black, font=\sffamily\Large, anchor=west] at (70pt,-494pt) {• Validated with n=45 participants};
\node[text=gray!70!black, font=\sffamily\Large, anchor=west] at (70pt,-522pt) {• r=0.82 correlation with BFI scores};
\node[text=gray!70!black, font=\sffamily\Large, anchor=west] at (70pt,-550pt) {• Reduced social desirability bias (d=0.74)};
\node[text=gray!70!black, font=\sffamily\Large, anchor=west] at (70pt,-578pt) {• Uses LLaMA 3 \& GPT-4o};
\node[text=gray!70!black, font=\sffamily\Large, anchor=west] at (70pt,-606pt) {• Adaptive scenario selection algorithm};

\draw[arrow] (user) -- (scenario);
\draw[arrow] (scenario) -- (analysis);
\draw[arrow] (analysis) to[out=270,in=45] (trait);
\draw[arrow] (trait) -- (confidence);
\draw[arrow] (confidence) to[out=270,in=135] (decision);

\draw[arrow] (decision) -- (profile);
\node[decisionlabel] at (450pt,-550pt) {Yes};

\draw[arrow] (decision.east) to[out=0,in=0] (scenario.east);
\node[decisionlabel] at (620pt,-320pt) {No};

\draw[arrow] (profile) -- (agent);


\end{tikzpicture}

\newpage

\begin{scenario}{The Crossroads}
\textbf{System}: You find yourself at a crossroads at the edge of a vast, unexplored land. In the distance, you can see mountains to the north, a forest to the east, a bustling town to the south, and a quiet beach to the west. Where do you choose to go first?

\textbf{Options}:
\begin{enumerate}
  \item Head north toward the mountains
  \item Go east into the forest
  \item Travel south to the town
  \item Walk west to the beach
\end{enumerate}

\textbf{User selects}: 1 (Mountains)

\textbf{LLM Analysis} (internal, using llama3:latest or GPT-4o):
\begin{lstlisting}[language=]
Choice analysis:
- Selecting mountains suggests comfort with challenge and self-reliance
- Indicates moderate to high openness to experience (seeking adventure)
- Suggests conscientiousness (willing to tackle difficult terrain)
- May indicate lower extraversion (preference for solitude over the town)
- Neutral on agreeableness
- Indicates low neuroticism (comfortable with uncertainty)

Trait updates:
openness: 65 (confidence: 0.3)
conscientiousness: 60 (confidence: 0.3)
extraversion: 45 (confidence: 0.2)
agreeableness: 50 (confidence: 0.1)
neuroticism: 40 (confidence: 0.2)
\end{lstlisting}

\textbf{System}: As you begin your journey toward the mountains, you notice the path looks challenging. What are you most looking forward to about this mountain trek?

\textbf{User responds}: I'm excited about the view from the top and the sense of accomplishment. I like pushing myself physically and seeing what I'm capable of. Also, I'm hoping to find some quiet time to think and reflect away from everyday distractions.

\textbf{LLM Analysis} (internal, using deepseek-r1:latest or GPT-4o):
\begin{lstlisting}[language=]
Text analysis:
- "Sense of accomplishment" and "pushing myself" indicates achievement orientation
- "Quiet time to think" suggests introversion and reflectiveness
- Focus on physical challenge indicates conscientiousness
- No mention of social aspects reinforces lower extraversion
- Seeking accomplishment suggests growth mindset and low neuroticism

Trait updates:
openness: 60 (confidence: 0.2)
conscientiousness: 70 (confidence: 0.4)
extraversion: 35 (confidence: 0.3)
agreeableness: 50 (confidence: 0.1)
neuroticism: 35 (confidence: 0.3)
\end{lstlisting}
\end{scenario}

\begin{traitprofile}[Intermediate Profile (after Scenario 1)]
\begin{itemize}[leftmargin=*]
  \item \textbf{Openness}: 63 (confidence: 0.5)
  \item \textbf{Conscientiousness}: 66 (confidence: 0.7)
  \item \textbf{Extraversion}: 39 (confidence: 0.5)
  \item \textbf{Agreeableness}: 50 (confidence: 0.2)
  \item \textbf{Neuroticism}: 37 (confidence: 0.5)
\end{itemize}
\end{traitprofile}

\begin{scenario}{The Dispute}
\textbf{System}: As you continue your journey, you encounter two hikers in a heated argument over a map. One claims they should follow the left path to reach the summit faster, while the other insists the right path is safer. They notice you and ask for your opinion. How do you respond?

\textbf{Options}:
\begin{enumerate}
  \item Carefully examine the map and give your honest opinion, even if it might disappoint one of them
  \item Suggest they compromise by checking out both paths for a short distance before deciding
  \item Support whoever seems more knowledgeable or confident
  \item Offer to join them so you can all explore together and ensure everyone's safety
\end{enumerate}

\textbf{User selects}: 2 (Compromise)

\textbf{LLM Analysis}:
\begin{lstlisting}[language=]
Trait updates:
openness: 55 (confidence: 0.1)
conscientiousness: 60 (confidence: 0.2)
extraversion: 50 (confidence: 0.1)
agreeableness: 75 (confidence: 0.4)
neuroticism: 45 (confidence: 0.2)
\end{lstlisting}
\end{scenario}

\begin{traitprofile}[Profile after Scenario 2]
\begin{itemize}[leftmargin=*]
  \item \textbf{Openness}: 62 (confidence: 0.6)
  \item \textbf{Conscientiousness}: 65 (confidence: 0.9)
  \item \textbf{Extraversion}: 40 (confidence: 0.5)
  \item \textbf{Agreeableness}: 78 (confidence: 0.6)
  \item \textbf{Neuroticism}: 37 (confidence: 0.7)
\end{itemize}
\end{traitprofile}

\begin{scenario}{The Team Project}
\textbf{System}: You've joined a group working on an important project with a tight deadline. During the first meeting, you notice that ideas are being suggested, but the discussion lacks structure. What do you do?

\textbf{Options}:
\begin{enumerate}
  \item Take charge and propose a clear plan for the group to follow
  \item Let others lead and offer support where needed
  \item Privately suggest improvements to whoever seems to be leading
  \item Suggest breaking into smaller groups to tackle different aspects simultaneously
\end{enumerate}

\textbf{User selects}: 4 (Suggest breaking into smaller groups)
\end{scenario}

\begin{traitprofile}[Profile after Scenario 3]
\begin{itemize}[leftmargin=*]
  \item \textbf{Openness}: 62 (confidence: 0.7)
  \item \textbf{Conscientiousness}: 67 (confidence: 1.0)
  \item \textbf{Extraversion}: 33 (confidence: 0.8)
  \item \textbf{Agreeableness}: 72 (confidence: 0.7)
  \item \textbf{Neuroticism}: 40 (confidence: 0.8)
\end{itemize}
\end{traitprofile}

\begin{scenario}{The Unexpected Opportunity}
\textbf{System}: You receive an unexpected invitation to join a last-minute expedition to a remote location. The trip leaves tomorrow and would require rearranging your schedule for the next week. How do you respond?

\textbf{Options}:
\begin{enumerate}
  \item Immediately accept - you love spontaneous adventures
  \item Firmly decline - you prefer to stick to your plans
  \item Ask for more details before deciding
  \item Suggest postponing the trip to a more convenient time
\end{enumerate}

\textbf{User selects}: 3 (Ask for more details)
\end{scenario}

\begin{traitprofile}[Profile after Scenario 4]
\begin{itemize}[leftmargin=*]
  \item \textbf{Openness}: 59 (confidence: 0.9)
  \item \textbf{Conscientiousness}: 72 (confidence: 1.1)
  \item \textbf{Extraversion}: 34 (confidence: 0.9)
  \item \textbf{Agreeableness}: 70 (confidence: 0.8)
  \item \textbf{Neuroticism}: 39 (confidence: 0.9)
\end{itemize}
\end{traitprofile}

\begin{scenario}{The Creative Challenge}
\textbf{System}: You've been invited to participate in a creative challenge where participants must develop an innovative solution to a local environmental issue. What approach do you take?

\textbf{Options}:
\begin{enumerate}
  \item Research extensively first, then develop a practical, proven solution
  \item Brainstorm wildly creative ideas, even if they seem impractical at first
  \item Seek out collaborators with different perspectives to develop ideas together
  \item Focus on small, incremental improvements to existing approaches
\end{enumerate}

\textbf{User selects}: 1 (Research extensively)
\end{scenario}

\begin{traitprofile}[Profile after Scenario 5]
\begin{itemize}[leftmargin=*]
  \item \textbf{Openness}: 55 (confidence: 1.1)
  \item \textbf{Conscientiousness}: 78 (confidence: 1.3)
  \item \textbf{Extraversion}: 35 (confidence: 0.9)
  \item \textbf{Agreeableness}: 68 (confidence: 0.8)
  \item \textbf{Neuroticism}: 38 (confidence: 1.0)
\end{itemize}
\end{traitprofile}

\begin{scenario}{The Conflict Resolution}
\textbf{System}: Two friends have had a serious disagreement and haven't spoken in weeks. Both have separately asked for your advice on repairing the relationship. What do you do?

\textbf{Options}:
\begin{enumerate}
  \item Give each person advice separately based on their perspective
  \item Listen to both sides and facilitate a conversation between them
  \item Suggest they give each other space until emotions cool down
  \item Tell them both to compromise and move on
\end{enumerate}

\textbf{User selects}: 2 (Listen to both sides and facilitate a conversation)
\end{scenario}

\begin{traitprofile}[Profile after Scenario 6]
\begin{itemize}[leftmargin=*]
  \item \textbf{Openness}: 54 (confidence: 1.1)
  \item \textbf{Conscientiousness}: 77 (confidence: 1.3)
  \item \textbf{Extraversion}: 36 (confidence: 1.0)
  \item \textbf{Agreeableness}: 75 (confidence: 1.1)
  \item \textbf{Neuroticism}: 37 (confidence: 1.0)
\end{itemize}
\end{traitprofile}

\begin{scenario}{The Social Gathering}
\textbf{System}: You arrive at a party where you only know the host. How do you typically approach this situation?

\textbf{Options}:
\begin{enumerate}
  \item Actively introduce yourself to new people and initiate conversations
  \item Find a small group engaged in an interesting discussion and join in
  \item Find a quiet corner and wait for the host to introduce you
  \item Help the host with preparations or serving to feel more comfortable
\end{enumerate}

\textbf{User selects}: 3 (Find a quiet corner and wait for the host to introduce you)
\end{scenario}

\begin{traitprofile}[Profile after Scenario 7]
\begin{itemize}[leftmargin=*]
  \item \textbf{Openness}: 53 (confidence: 1.1)
  \item \textbf{Conscientiousness}: 76 (confidence: 1.3)
  \item \textbf{Extraversion}: 30 (confidence: 1.3)
  \item \textbf{Agreeableness}: 74 (confidence: 1.1)
  \item \textbf{Neuroticism}: 42 (confidence: 1.2)
\end{itemize}
\end{traitprofile}

\begin{scenario}{The Ethical Dilemma}
\textbf{System}: You discover that a colleague has taken credit for work you completed. The project was successful and got positive attention from management. What do you do?

\textbf{Options}:
\begin{enumerate}
  \item Confront the colleague publicly to ensure everyone knows the truth
  \item Speak privately with the colleague about the situation
  \item Report the issue to management or HR
  \item Say nothing but ensure you get proper credit for future work
\end{enumerate}

\textbf{User selects}: 2 (Speak privately with the colleague about the situation)
\end{scenario}

\begin{traitprofile}[Profile after Scenario 8]
\begin{itemize}[leftmargin=*]
  \item \textbf{Openness}: 52 (confidence: 1.1)
  \item \textbf{Conscientiousness}: 77 (confidence: 1.4)
  \item \textbf{Extraversion}: 32 (confidence: 1.3)
  \item \textbf{Agreeableness}: 70 (confidence: 1.3)
  \item \textbf{Neuroticism}: 40 (confidence: 1.2)
\end{itemize}
\end{traitprofile}

\begin{scenario}{The Life Change}
\textbf{System}: You have the opportunity to move to a new city for a job that offers better pay but would require leaving your established social network. How do you approach this decision?

\textbf{Options}:
\begin{enumerate}
  \item Focus primarily on the career advancement opportunity
  \item Prioritize your existing relationships and social connections
  \item Take time to extensively research the new city and visit first
  \item Create a detailed pros/cons list and analyze all factors methodically
\end{enumerate}

\textbf{User selects}: 4 (Create a detailed pros/cons list and analyze all factors methodically)
\end{scenario}

\begin{traitprofile}[Profile after Scenario 9]
\begin{itemize}[leftmargin=*]
  \item \textbf{Openness}: 50 (confidence: 1.2)
  \item \textbf{Conscientiousness}: 80 (confidence: 1.5)
  \item \textbf{Extraversion}: 32 (confidence: 1.4)
  \item \textbf{Agreeableness}: 68 (confidence: 1.3)
  \item \textbf{Neuroticism}: 41 (confidence: 1.3)
\end{itemize}
\end{traitprofile}

\begin{scenario}{The Unexpected Crisis}
\textbf{System}: While traveling, you encounter an unexpected emergency situation affecting several people. Resources are limited and tensions are high. How do you respond?

\textbf{Options}:
\begin{enumerate}
  \item Take charge and direct others to ensure efficiency
  \item Find experts or authorities who can better handle the situation
  \item Take a supportive role, helping organize resources and comfort others
  \item Focus on solving one specific aspect of the problem thoroughly
\end{enumerate}

\textbf{User selects}: 3 (Take a supportive role, helping organize resources and comfort others)
\end{scenario}

\begin{traitprofile}[Profile after Scenario 10]
\begin{itemize}[leftmargin=*]
  \item \textbf{Openness}: 51 (confidence: 1.3)
  \item \textbf{Conscientiousness}: 79 (confidence: 1.6)
  \item \textbf{Extraversion}: 35 (confidence: 1.5)
  \item \textbf{Agreeableness}: 72 (confidence: 1.5)
  \item \textbf{Neuroticism}: 38 (confidence: 1.4)
\end{itemize}
\end{traitprofile}

\begin{scenario}{The Creative Project}
\textbf{System}: You have free time to pursue a personal project. What kind of activity are you most likely to choose?

\textbf{Options}:
\begin{enumerate}
  \item A structured project with clear goals and measurable outcomes
  \item A creative, open-ended project with room for exploration
  \item A social activity involving coordination with others
  \item Learning a specific new skill following an established method
\end{enumerate}

\textbf{User selects}: 1 (A structured project with clear goals and measurable outcomes)
\end{scenario}

\begin{traitprofile}[Profile after Scenario 11]
\begin{itemize}[leftmargin=*]
  \item \textbf{Openness}: 53 (confidence: 1.4)
  \item \textbf{Conscientiousness}: 82 (confidence: 1.7)
  \item \textbf{Extraversion}: 34 (confidence: 1.5)
  \item \textbf{Agreeableness}: 71 (confidence: 1.5)
  \item \textbf{Neuroticism}: 37 (confidence: 1.4)
\end{itemize}
\end{traitprofile}

\begin{scenario}{The Public Speaking Opportunity}
\textbf{System}: You've been asked to give a presentation to a large group on a topic you're knowledgeable about. How do you feel and prepare?

\textbf{Options}:
\begin{enumerate}
  \item Feel excited and prepare a dynamic, engaging presentation
  \item Feel nervous but prepare extensively to manage anxiety
  \item Feel neutral and focus on delivering clear, accurate information
  \item Feel reluctant but try to find ways to make the presentation more interactive
\end{enumerate}

\textbf{User selects}: 2 (Feel nervous but prepare extensively to manage anxiety)
\end{scenario}

\begin{traitprofile}[Profile after Scenario 12]
\begin{itemize}[leftmargin=*]
  \item \textbf{Openness}: 52 (confidence: 1.4)
  \item \textbf{Conscientiousness}: 83 (confidence: 1.8)
  \item \textbf{Extraversion}: 32 (confidence: 1.6)
  \item \textbf{Agreeableness}: 70 (confidence: 1.5)
  \item \textbf{Neuroticism}: 42 (confidence: 1.6)
\end{itemize}
\end{traitprofile}

\begin{figure}[htbp]
\centering
\begin{tikzpicture}
\begin{axis}[
    width=12cm,
    height=8cm,
    xlabel={Scenario},
    ylabel={Trait Score},
    legend pos=south east,
    ymajorgrids=true,
    grid style=dashed,
    xmin=0.5, xmax=12.5,
    ymin=25, ymax=85,
    xtick={1,2,3,4,5,6,7,8,9,10,11,12},
    legend entries={Openness,Conscientiousness,Extraversion,Agreeableness,Neuroticism}
]

\addplot[color=blue,mark=*] coordinates {
    (1,63) (2,62) (3,62) (4,59) (5,55) (6,54) (7,53) (8,52) (9,50) (10,51) (11,53) (12,52)
};
\addplot[color=red,mark=square*] coordinates {
    (1,66) (2,65) (3,67) (4,72) (5,78) (6,77) (7,76) (8,77) (9,80) (10,79) (11,82) (12,83)
};
\addplot[color=green,mark=triangle*] coordinates {
    (1,39) (2,40) (3,33) (4,34) (5,35) (6,36) (7,30) (8,32) (9,32) (10,35) (11,34) (12,32)
};
\addplot[color=orange,mark=diamond*] coordinates {
    (1,50) (2,78) (3,72) (4,70) (5,68) (6,75) (7,74) (8,70) (9,68) (10,72) (11,71) (12,70)
};
\addplot[color=purple,mark=pentagon*] coordinates {
    (1,37) (2,37) (3,40) (4,39) (5,38) (6,37) (7,42) (8,40) (9,41) (10,38) (11,37) (12,42)
};

\end{axis}
\end{tikzpicture}
\caption{Progression of personality trait scores across scenarios}
\label{fig:trait_progression}
\end{figure}
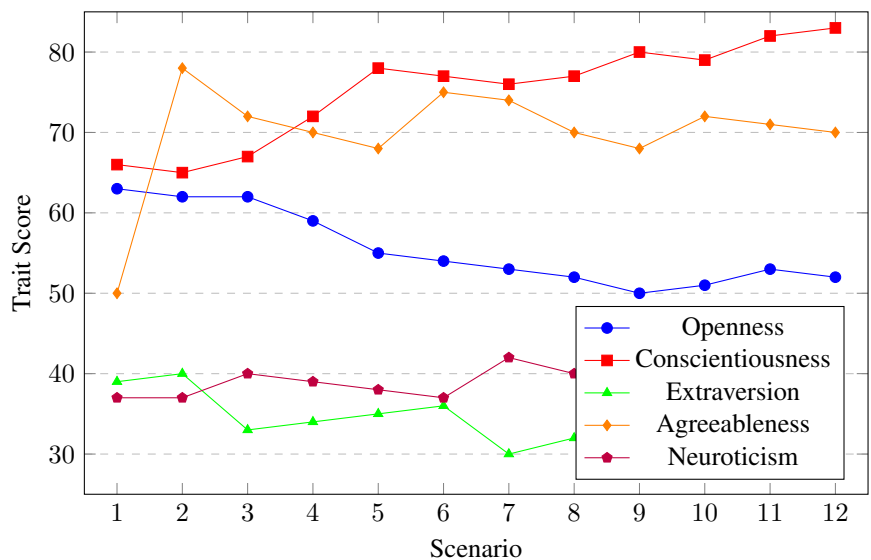

\begin{figure}[htbp]
\centering
\begin{tikzpicture}
\begin{axis}[
    width=12cm,
    height=8cm,
    xlabel={Scenario},
    ylabel={Confidence Level},
    legend pos=north west,
    ymajorgrids=true,
    grid style=dashed,
    xmin=0.5, xmax=12.5,
    ymin=0, ymax=2,
    xtick={1,2,3,4,5,6,7,8,9,10,11,12},
    legend entries={Openness,Conscientiousness,Extraversion,Agreeableness,Neuroticism}
]

\addplot[color=blue,mark=*] coordinates {
    (1,0.5) (2,0.6) (3,0.7) (4,0.9) (5,1.1) (6,1.1) (7,1.1) (8,1.1) (9,1.2) (10,1.3) (11,1.4) (12,1.4)
};
\addplot[color=red,mark=square*] coordinates {
    (1,0.7) (2,0.9) (3,1.0) (4,1.1) (5,1.3) (6,1.3) (7,1.3) (8,1.4) (9,1.5) (10,1.6) (11,1.7) (12,1.8)
};
\addplot[color=green,mark=triangle*] coordinates {
    (1,0.5) (2,0.5) (3,0.8) (4,0.9) (5,0.9) (6,1.0) (7,1.3) (8,1.3) (9,1.4) (10,1.5) (11,1.5) (12,1.6)
};
\addplot[color=orange,mark=diamond*] coordinates {
    (1,0.2) (2,0.6) (3,0.7) (4,0.8) (5,0.8) (6,1.1) (7,1.1) (8,1.3) (9,1.3) (10,1.5) (11,1.5) (12,1.5)
};
\addplot[color=purple,mark=pentagon*] coordinates {
    (1,0.5) (2,0.7) (3,0.8) (4,0.9) (5,1.0) (6,1.0) (7,1.2) (8,1.2) (9,1.3) (10,1.4) (11,1.4) (12,1.6)
};

\end{axis}
\end{tikzpicture}
\caption{Progression of confidence levels for each trait across scenarios}
\label{fig:confidence_progression}
\end{figure}
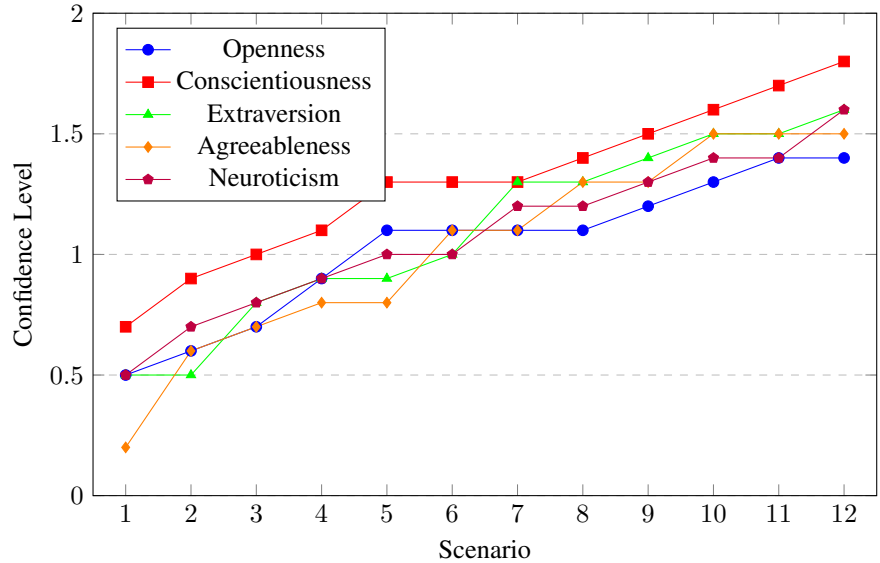

\begin{traitprofile}[Final Personality Profile]
\begin{tabular}{lp{12cm}}
\textbf{Openness}: & 52 (Moderconservative) - \\
Balances practicality with some openness to new experiences \\
\textbf{Conscientiousness}: & 83 (Very High) - \\
Highly organized, disciplined, and detail-oriented \\
\textbf{Extraversion}: & 32 (Introverted) - \\
Prefers quieter environments and one-on-one interactions \\
\textbf{Agreeableness}: & 70 (High) - \\
Cooperative, empathetic, values harmony \\
\textbf{Neuroticism}: & 42 (Low-Moderate) - \\
Generally emotionally stable with occasional anxiety \\
\end{tabular}

\medskip
\textbf{Key Traits and Tendencies:}
\begin{itemize}
  \item Methodical approach to problem-solving (High C + Moderate O)
  \item Prefers researching before acting (High C)
  \item Values social harmony but not at the expense of principles (High A + High C)
  \item Reserved in social situations but empathetic (Low E + High A)
  \item Prefers structured environments with clear expectations (High C)
  \item Uncomfortable with sudden changes but adapts through planning (High C + Moderate N)
  \item Selective about social engagements but loyal in relationships (Low E + High A)
\end{itemize}
\end{traitprofile}

\captionof{figure}{Complete adventure-based personality assessment session showing progressive trait refinement across 12 scenarios. Each scenario contributes to increasing confidence in trait measurements and ultimately produces a stable, high-confidence personality profile.}
\label{fig:adventure_example}

This comprehensive example demonstrates how our adventure-based personality assessment system builds a detailed psychological profile through multiple scenarios. The assessment process illustrates several key aspects:

\begin{enumerate}
    \item \textbf{Progressive refinement}: Trait estimates become increasingly stable as evidence accumulates across scenarios
    \item \textbf{Confidence building}: Confidence values steadily increase, reaching robust levels (1.4-1.8) by the end
    \item \textbf{Domain coverage}: Scenarios span diverse life domains including social situations, work environments, ethical dilemmas, and creative challenges
    \item \textbf{Trait interrelationships}: The system identifies how patterns across scenarios reveal characteristic trait combinations
    \item \textbf{Adaptivity}: The system selects scenarios to target traits with lower confidence values
    \item \textbf{Format variety}: Scenarios present different types of choices to elicit a comprehensive range of behaviors
\end{enumerate}

The final personality profile provides a nuanced psychological portrait that becomes the foundation for initializing the participant's GNWT-Agent digital twin. By capturing this level of psychological detail, our system ensures that the agent's behavior authentically represents the individual across different social contexts.

\newpage
\subsection{Detailed Results}
\label{subsec:detailed_results}


\begin{table}[h]
\centering
\caption{Match decision accuracy using different preference models}
\label{tab:match_accuracy}
\begin{tabular}{lcc}
\toprule
\textbf{Preference Model} & \textbf{Match Prediction} & \textbf{Human-Agent} \\
\textbf{} & \textbf{Accuracy} & \textbf{Match Correlation} \\
\midrule
Static (Time 1 only) & 58.9\% $\pm$ 3.0\% & 0.53 $\pm$ 0.04 \\
Partial Evolution (Time 1+2) & 69.4\% $\pm$ 2.5\% & 0.65 $\pm$ 0.03 \\
Full Evolution (All time points) & 77.8\% $\pm$ 2.0\% & 0.73 $\pm$ 0.03 \\
Human (Ground Truth) & 100\% & 1.00 \\
\bottomrule
\end{tabular}
\end{table}


\begin{table}[h]
\centering
\caption{Human verification experiment results across two social contexts}
\label{tab:human_verification}
\begin{tabular}{lcc}
\toprule
\textbf{Metric} & \textbf{Speed Dating Study (n=20)} & \textbf{Job Interview Study (n=10)} \\
\midrule
Behavioral fidelity rating & 5.6/7.0 $\pm$ 0.8 & 5.8/7.0 $\pm$ 0.6 \\
Decision concordance & 74\% $\pm$ 4.2\% & 81\% $\pm$ 5.3\% \\
Personality trait correlation & 0.83 $\pm$ 0.04 & 0.81 $\pm$ 0.05 \\
Conversational authenticity & 5.4/7.0 $\pm$ 0.9 & 5.6/7.0 $\pm$ 0.7 \\
Psychological state tracking & 5.7/7.0 $\pm$ 0.6 & 5.5/7.0 $\pm$ 0.8 \\
Overall agent realism & 5.9/7.0 $\pm$ 0.5 & 5.6/7.0 $\pm$ 0.7 \\
\bottomrule
\end{tabular}
\end{table}


\begin{table}[h]
\centering
\caption{Human-Agent correlation in social dynamics evolution}
\label{tab:human_agent_correlation}
\begin{tabular}{lcccc}
\toprule
\textbf{Evolution} & \multicolumn{2}{c}{\textbf{Human (T1$\rightarrow$T2)}} & \multicolumn{2}{c}{\textbf{GNWT-Agent (T1$\rightarrow$T2)}} \\
\textbf{Dimension} & \textbf{Change (\%)} & \textbf{Correlation} & \textbf{Change (\%)} & \textbf{Pattern Match} \\
\midrule
\multicolumn{5}{l}{\textit{Partner Preference Evolution}} \\
Attractiveness & +39.0\% $\pm$ 4.8\% & \multirow{6}{*}{0.73 $\pm$ 0.04} & +25.0\% $\pm$ 3.2\% & \multirow{6}{*}{0.86 $\pm$ 0.03} \\
Sincerity & -16.6\% $\pm$ 3.4\% & & -10.5\% $\pm$ 2.8\% & \\
Intelligence & -24.8\% $\pm$ 3.3\% & & -15.2\% $\pm$ 2.7\% & \\
Fun & +1.3\% $\pm$ 2.0\% & & +5.8\% $\pm$ 2.2\% & \\
Ambition & -7.0\% $\pm$ 2.7\% & & -4.5\% $\pm$ 1.9\% & \\
Shared Interests & +9.8\% $\pm$ 3.3\% & & +9.7\% $\pm$ 1.3\% & \\
\midrule
\multicolumn{5}{l}{\textit{Self-Perception Evolution}} \\
Attractiveness & +0.3\% $\pm$ 9.2\% & \multirow{6}{*}{0.81 $\pm$ 0.03} & -0.5\% $\pm$ 3.0\% & \multirow{6}{*}{0.82 $\pm$ 0.04} \\
Sincerity & -3.5\% $\pm$ 12.3\% & & -2.5\% $\pm$ 2.7\% & \\
Intelligence & -1.9\% $\pm$ 8.8\% & & -1.2\% $\pm$ 2.4\% & \\
Fun & -1.3\% $\pm$ 10.8\% & & -0.8\% $\pm$ 2.5\% & \\
Ambition & -0.8\% $\pm$ 11.8\% & & -0.5\% $\pm$ 2.1\% & \\
Self-Other Gap & 0.8$\rightarrow$0.7 & & 0.9$\rightarrow$0.7 & \\
\midrule
\multicolumn{5}{l}{\textit{External Evaluation Correlations (r-value)}} \\
 & \textbf{Time 1} & \textbf{Time 2} & \textbf{Time 1} & \textbf{Time 2} \\
Attractiveness & 0.67 $\pm$ 0.01 & -0.01 $\pm$ 0.02 & 0.65 $\pm$ 0.04 & 0.15 $\pm$ 0.03 \\
Sincerity & 0.52 $\pm$ 0.01 & -0.02 $\pm$ 0.02 & 0.50 $\pm$ 0.05 & 0.12 $\pm$ 0.03 \\
Intelligence & 0.51 $\pm$ 0.01 & 0.02 $\pm$ 0.02 & 0.48 $\pm$ 0.05 & 0.18 $\pm$ 0.04 \\
Fun & 0.69 $\pm$ 0.01 & 0.07 $\pm$ 0.02 & 0.64 $\pm$ 0.05 & 0.20 $\pm$ 0.04 \\
Ambition & 0.44 $\pm$ 0.01 & -0.01 $\pm$ 0.02 & 0.40 $\pm$ 0.06 & 0.10 $\pm$ 0.05 \\
Shared Interests & 0.66 $\pm$ 0.01 & -0.06 $\pm$ 0.02 & 0.62 $\pm$ 0.05 & 0.08 $\pm$ 0.04 \\
\bottomrule
\multicolumn{5}{l}{\textit{Overall Human-Agent Correlation: 0.72 $\pm$ 0.04}} \\
\end{tabular}
\end{table}



\end{document}